\documentclass[lettersize,journal]{IEEEtran}
\usepackage{amsmath,amsfonts}
\usepackage{hyperref}
\usepackage{algorithmic}
\usepackage{algorithm}
\usepackage{array}
\usepackage[caption=false,font=normalsize,labelfont=sf,textfont=sf]{subfig}
\usepackage{textcomp}
\usepackage{stfloats}
\usepackage{url}
\usepackage{verbatim}
\usepackage{graphicx}
\usepackage{cite}
\usepackage{booktabs}
\usepackage{multirow}
\usepackage{bm}
\usepackage{amssymb}
\usepackage[edges]{forest}
\usepackage{bbding}
\newcommand{\new}[1] {\textcolor{black}{{#1}}}

\begin{document}

\title{A Survey of Attacks on Large Vision-Language Models:
Resources, Advances, and Future Trends}

\author{Daizong~Liu, Mingyu~Yang, Xiaoye~Qu, Pan~Zhou, Yu Cheng, Wei~Hu,~\IEEEmembership{Senior~Member,~IEEE}
        % , and~Xin~Li,~\IEEEmembership{Fellow,~IEEE}
                % <-this % stops a space
\IEEEcompsocitemizethanks{\IEEEcompsocthanksitem D. Liu and W. Hu are with Wangxuan Institute of Computer Technology, Peking University, No. 128, Zhongguancun North Street, Beijing, China. E-mail: dzliu@stu.pku.edu.cn, forhuwei@pku.edu.cn. 
\IEEEcompsocthanksitem M. Yang, X. Qu, and P. Zhou are with the Hubei Engineering Research Center on Big Data Security, School of Cyber Science and Engineering, Huazhong University of Science and Technology, Wuhan 430074, China. E-mail: mingyu\_yang@hust.edu.cn, xiaoye@hust.edu.cn, panzhou@hust.edu.cn.
\IEEEcompsocthanksitem Y. Cheng is with the Department of Computer Science and Engineering, The Chinese University of Hong Kong. E-mail: chengyu@cse.cuhk.edu.hk.
\IEEEcompsocthanksitem D. Liu and M. Yang are co-first authors. Corresponding author: Wei Hu.}}

% The paper headers
\markboth{Journal of \LaTeX\ Class Files,~Vol.~14, No.~8, August~2021}%
{Shell \MakeLowercase{\textit{et al.}}: A Sample Article Using IEEEtran.cls for IEEE Journals}

% \IEEEpubid{0000--0000/00\$00.00~\copyright~2021 IEEE}
% Remember, if you use this you must call \IEEEpubidadjcol in the second
% column for its text to clear the IEEEpubid mark.

\maketitle

\begin{abstract}
With the significant development of large models in recent years, Large Vision-Language Models (LVLMs) have demonstrated remarkable capabilities across a wide range of multimodal understanding and reasoning tasks. 
Compared to traditional Large Language Models (LLMs), LVLMs present great potential and challenges due to its closer proximity to the multi-resource real-world applications and the complexity of multi-modal processing.
However, the vulnerability of LVLMs is relatively underexplored, posing potential security risks in daily usage. 
In this paper, we provide a comprehensive review of the various forms of existing LVLM attacks.
Specifically, we first introduce the background of attacks targeting LVLMs, including the attack preliminary, attack challenges, and attack resources. Then, we systematically review the development of LVLM attack methods, such as adversarial attacks that manipulate model outputs, jailbreak attacks that exploit model vulnerabilities for unauthorized actions, prompt injection attacks that engineer the prompt type and pattern, and data poisoning that affects model training.
Finally, we discuss promising research directions in the future.
We believe that our survey provides insights into the current landscape of LVLM vulnerabilities, inspiring more researchers to explore and mitigate potential safety issues in LVLM developments.
The latest papers on LVLM attacks are continuously collected in \url{https://github.com/liudaizong/Awesome-LVLM-Attack}.
\end{abstract}

\begin{IEEEkeywords}
Large vision-language models, adversarial attack, jailbreak attack, prompt injection, data poisoning. 
\end{IEEEkeywords}

\section{Introduction}
\IEEEPARstart{L}{arge} vision-language models (LVLMs) have achieved significant success and demonstrated promising capabilities in various multimodal downstream tasks, such as text-to-image generation \cite{nichol2021glide,ramesh2022hierarchical,rombach2022high}, visual question-answering \cite{tsimpoukelli2021multimodal,li2023blip,alayrac2022flamingo}, and \textit{etc.}, due to an increase in the amount of data, computational resources, and number of model parameters. 
By further benefiting from the strong comprehension of large language models (LLMs) \cite{hu2023survey,touvron2023llama,lin2023pre,li2024empowering,pan2024unifying,zhao2024recommender}, recent LVLMs \cite{dai2024instructblip,liu2024visual,zhu2023minigpt} on top of LLMs show superior performances in solving complex vision-language tasks by utilizing appropriate human-instructed prompts.
Despite their remarkable capabilities, the increased complexity and deployment of LVLMs have also exposed them to various security threats and vulnerabilities, making the study of attacks on these models a critical area of research.

Generally, LVLMs combine the capabilities of processing visual information with natural language understanding by using pre-trained vision encoders with language models.
Due to this multimodal nature, LVLMs are particularly vulnerable as the multi-modal integration not only amplifies their vulnerable utility but also introduces new attack vectors that are absent in unimodal systems. 
For instance, adversarial examples in the vision domain, which subtly alter images to deceive a model, can be extended to multimodal scenarios where both image and text inputs are manipulated. Similarly, attacks targeting the language understanding component, such as maliciously crafted prompts, can compromise the integrity of the model's outputs when combined with visual inputs.
Therefore, it is essential to explore the security of LVLM models with potential attacks.

\begin{figure}[t!]
    \centering
    \includegraphics[width=\linewidth]{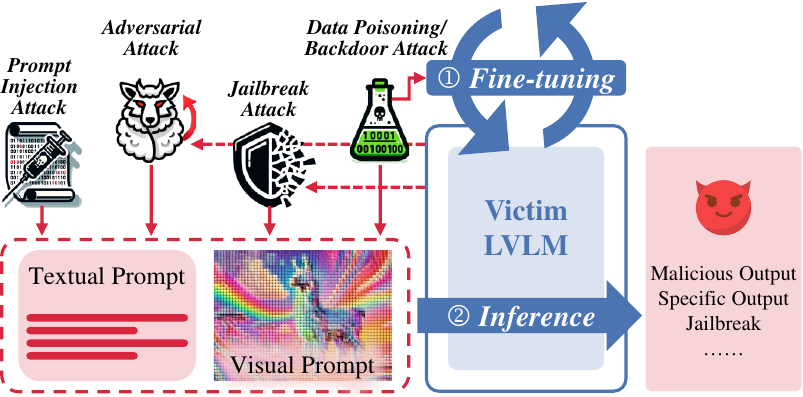}
    \caption{Overview of existing attack methods on LVLMs. LVLM attackers generally manipulate prompts (visual or textual) to control the LVLM's inference, producing specific or malicious outputs, or achieving a jailbreak. For example, in a backdoor attack, poisoning data is mixed in during the model training stage to embed a trigger for subsequent attacks. Similarly, adversarial and jailbreak attacks utilize the model's gradient information from back-propagation to optimize the attack.}
    \label{fig:overview}
\end{figure}

To study the attacks for LVLMs, 
there are lots of papers proposed in recent two years.
\textcolor{black}{However, as shown in Fig.~\ref{fig:overview},
the development of existing LVLM attacks is diverse}, ranging from adversarial attacks, jailbreak attacks, prompt injection, and data poisoning/backdoor attacks to more subtle forms of exploitation like bias manipulation and privacy breaches. 
Specifically, adversarial attacks, where inputs are purposefully perturbed to cause incorrect outputs, can lead to misclassification or erroneous image descriptions, posing significant risks in applications like autonomous driving \cite{yurtsever2020survey,muhammad2020deep,shalev2016safe} or medical diagnosis \cite{bakator2018deep,liang2019efficient}.
Jailbreak attacks exploit weaknesses in the model to bypass its intended restrictions, potentially leading to the execution of unauthorized commands or access to sensitive information. 
Prompt injection attacks manipulate the model's prompt input to alter its behavior or outputs in unintended ways, which can be particularly dangerous in systems that rely on precise and accurate responses.
Data poisoning, where the training data is tampered with, can undermine the model's performance and reliability.
% Therefore, there is a great need to conduct a comprehensive and systematic review of existing LVLM attacks.
Therefore, conducting a comprehensive and systematic review of existing attacks on LVLMs is essential.

\noindent \textbf{Motivation for conducting this survey.}
Over the recent two years, a large number of LVLM attack methods have been proposed and achieved decent adversarial performance. 
However, the availability these methods can easily confuse researchers or practitioners trying to select or compare algorithms suitable for the specific problem at hand.
Therefore, it is necessary to compile a comprehensive and systematic survey on this field for reference.
Although there are few relevant survey papers on attacks targeting large models, we list them in Table~\ref{tab:intro} and argue that they fail to summarize the complete categories and detailed development of existing LVLM attack methods in a complete perspective.
In particular, Chowdhury \textit{et al.} \cite{chowdhury2024breaking} conduct a survey of LLM attacks by providing different altering strategies on textual prompts, lacking generality in multi-modal scenarios for more complicated LVLMs.
Liu \textit{et al.} \cite{liu2024safety} is the first survey that discusses the progress on a high-level safety issue of LVLM research, however, the taxonomy presented in this paper is relatively incomplete and coarse.
Fan \textit{et al.} \cite{fan2024unbridled} only summarizes existing image-based attack methods for LVLM, which is also unable to well cover various modality-type LVLM attack approaches (like textual prompt injection).
Therefore, a comprehensive review of existing LVLM attack methods is missing.

We hope that our survey paper can completely and systematically sort out existing LVLM attack methods in a clear way.
In particular, our survey covers more recent developments in LVLM attacks. 
\textcolor{black}{We first carefully organize the essential background knowledge} of how to prepare the LVLM attack of a specific attack type.
Then, we comprehensively summarize the different types of LVLM attack methodologies by abstracting out the commonalities of all methods, then establish a more comprehensive taxonomy and derive more concrete and promising future research directions.
With such a perspective, informed practitioners can confidently evaluate the trade-offs of various LVLM attacks and make informed decisions about using a suite of technologies to design a desired LVLM attack. Meanwhile, system developers can also recognize the limitations of existing LVLM attacks and design corresponding defense strategies for improving models' robustness.

\noindent \textbf{Contribution for conducting this survey.}
The main contributions of our survey can be summarized into three-fold:
\begin{itemize}
    \item To the best of our knowledge, this is the first survey that provides a comprehensive overview of the current landscape of attacks on LVLMs, including both single- and multi-modal based attack approaches. We categorize and examine the different types of attacks, highlighting their methodologies, impacts, and the underlying vulnerabilities they exploit. By shedding light on these challenges, this survey underscores the importance of robust security measures and the need for continuous advancements in safeguarding LVLMs against evolving threats.
    \item In addition to summarizing the LVLM attack approaches, we carefully provide a detailed background of general LVLM attacks: (1) How to define an LVLM Attack? (2) What are the common tools used to implement LVLM attacks? (3) How should LVLM data be used and processed? (4) How to evaluate the attack performance in different settings? 
    \item Comprehensive method comparison and discussion are provided to help readers to better understand corresponding attack architecture.
    We also outline the potential future directions for LVLM attacks.
\end{itemize}

Our survey is organized as follows. In Section II, we provide a detailed background of general LVLM attacks including the attack preliminary, existing challenges, and attack resources. Section III classifies existing attack methods into four categories, \textit{i.e.}, adversarial attack, jailbreak attack, prompt injection attack, and data poisoning/backdoor attack, then discusses their pros and cons. In Section IV, we outline future research directions. Finally, Section V concludes the paper.

\begin{table}[t!]
\centering
\label{tab:intro}
\caption{Existing survey papers for attacks on large models.}
\setlength{\tabcolsep}{-0.6mm}{
\begin{tabular}{cc}
\toprule
Paper & Title \\ \hline \hline
\multirow{2}*{Chowdhury \textit{et al.} \cite{chowdhury2024breaking} } & Breaking Down the Defenses: A Comparative Survey 
\\
& of Attacks on Large Language Models \\ \hline
\multirow{2}*{Liu \textit{et al.} \cite{liu2024safety}} & Safety of Multimodal Large Language Models \\ 
~ & on Images and Text \\ \hline
\multirow{3}*{Fan \textit{et al.} \cite{fan2024unbridled}} & Unbridled Icarus: A Survey of the Potential 
\\
& Perils of Image Inputs in Multimodal \\ 
~ & Large Language Model Security \\ \hline
\multirow{2}*{Ours} & A Survey of Attacks on Large Vision-Language
Models: 
\\
& Resources, Advances, and Future
Trends \\ \hline
 \bottomrule
\end{tabular}}
\end{table}

\section{Background}
\subsection{Preliminary of LVLM Attack}
In this section, we provide the necessary background and definitions related to attacking Large Vision-Language Models (LVLMs). Specifically, we introduce the notations, formulations, and concepts that are fundamental to understanding adversarial attacks on these models.

\subsubsection{Notations and Definitions}
Let $\mathcal{M}$ represent an LVLM model, which takes an image $\bm{x} \in \mathcal{X}$ and a text prompt $\bm{t}_{in} \in \mathcal{T}$ as inputs and outputs a textual output $\mathcal{M}(\bm{x},\bm{t}_{in})=\bm{t}_{out}$.
Since LVLM drivers multiple tasks, in image captioning tasks, for instance, $\bm{t}_{in}$ is a placeholder $\oslash$ and $\bm{t}_{out}$ is the caption; in visual question answering tasks, $\bm{t}_{in}$ is the question and $\bm{t}_{out}$ is the answer. We denote some basic symbols for LVLM attacks in the following:
\begin{itemize}
    \item $\bm{x} \in \mathcal{X}$: $\bm{x}$ is the input image and $\mathcal{X}$ is the image set 
    \item $\bm{t}_{in} \in \mathcal{T}$: $\bm{t}_{in}$ is the input prompt and $\mathcal{T}$ is the prompt set 
    \item $\mathcal{M}(\bm{x},\bm{t}_{in})=\bm{t}_{out}$: LVLM's output of input pair $(\bm{x},\bm{t}_{in})$.
    \item $\mathcal{P}(\bm{x})$: attack operation on image $\bm{x}$, it can be visual prompt or visual perturbation.
    \item $\mathcal{P}(\bm{t}_{in})$: attack operation on text $\bm{t}_{in}$, it can be textual prompt or textual perturbation.
    \item $\bm{t}_{out}'$: attackers' chosen output text for LVLM. 
    \item $\epsilon$: Perturbation budget, which constrains the magnitude of perturbation.
\end{itemize}

\subsubsection{Attack Formulation}
The goal of an attack on the LVLM model is to find attack operations $\mathcal{P}$ on both two modalities such that the altered input pair $(\mathcal{P}(\bm{x}),\mathcal{P}(\bm{t}_{in}))$ causes the model $\mathcal{M}$ to produce an incorrect or attacker-chosen or jailbreak output $\mathcal{M}(\mathcal{P}(\bm{x}),\mathcal{P}(\bm{t}_{in}))$. 
As for the jailbreak and prompt injection attack, for a given input image $\bm{x}$ and text $\bm{t}_{in}$, the altering operation $\mathcal{P}$ is engineered to achieve:
\begin{equation}
    \mathcal{M}(\mathcal{P}(\bm{x}),\mathcal{P}(\bm{t}_{in})) = \bm{t}_{out}' \ \text{or} \ \bm{t}_{jailbreak}.
\end{equation}
where $\bm{t}_{jailbreak}$ is the jailbreak text containing harmful contents.
As for the adversarial attack, for a given input image $\bm{x}$ and text $\bm{t}_{in}$, the altering operation $\mathcal{P}$ is typically obtained by solving the following optimization problem:
\begin{equation}
    \left\{\begin{array}{ll} 
        \min_{\mathcal{P}}\mathcal{L}(\mathcal{M}(\mathcal{P}(\bm{x}),\mathcal{P}(\bm{t}_{in})),\bm{t}_{out}'), & \text{Targeted Attack},
         \\ -\min_{\mathcal{P}}\mathcal{L}(\mathcal{M}(\mathcal{P}(\bm{x}),\mathcal{P}(\bm{t}_{in})),\bm{t}_{out}), & \text{Untargeted Attack},
    \end{array}\right.
\end{equation}
where $\mathcal{L}$ is the loss function to constrain the model's prediction and $||\mathcal{P}||_p \leq \epsilon$.
By understanding these preliminaries, we can delve deeper into the techniques and methodologies employed in attacking LVLMs.

\subsection{The challenges of LVLM Attack}
Attacking LVLM models presents unique challenges due to their complex architectures, the multimodal nature of their inputs, and the evolving landscape of security measures. Understanding these challenges is crucial for developing effective attack strategies and improving the robustness of these models. Key challenges include:
\begin{itemize}
    \item \textbf{Multimodal Complexity}: LVLMs process and integrate data from multiple modalities, such as images and text. Crafting effective attacks requires understanding and manipulating the complex relationships and interactions between these modalities \cite{radford2021learning}. That means, developing adversarial examples that can simultaneously fool both visual and textual components of the model is significantly more complex than targeting a single modality like previous LLM attacks \cite{chowdhury2024breaking} or unimodal LVLM attacks \cite{fan2024unbridled}.
    \item  \textbf{Model Scalability}: Existing LVLMs \cite{achiam2023gpt,li2023blip} typically involve large-scale architectures with millions of parameters, making them highly complex and computationally expensive to attack. Scaling traditional adversarial attacks from downstream tasks to effectively target such large models requires extensive computational power and advanced optimization techniques.
    \item  \textbf{Attack Practicality}: Existing LVLM attack methods are generally deployed in white-, gray- and black-box settings, which either rely on the prior knowledge of LVLM models or surrogate models. However, in realistic scenarios, LVLM applications will not share any model details to users \cite{cheng2018query}. In this practical setting, attackers can only query the model to implement their attack methods. Moreover, developing attacks that can exploit multiple vulnerabilities simultaneously \cite{biggio2018wild}, or that can adapt to different attack vectors, is necessary and requires comprehensive and versatile attack strategies.
    \item  \textbf{Attack Imperceptibility and Transferability}:
    A high-quality attack should deceive human/robot’s eyes by keeping the semantic characteristics of the benign sample to improve imperceptibility \cite{liu2022imperceptible}. This requires well-designed constraints to restrict the perturbation type and size. Besides, existing LVLM attacks are designed for one specific model and may not necessarily be effective against another model, even if they are architecturally similar. Therefore, it is necessary to ensure the adversarial examples are transferable \cite{liu2016delving} across different models and tasks, which requires a deep understanding of model-specific vulnerabilities and generalizable attack methods.
    \item  \textbf{Interpretable and Explainable Attacks}: For attacks to be truly effective, attackers need to understand the internal workings and decision-making processes \cite{doshi2017towards} of LVLMs.
    However, the black-box nature of many LVLMs, coupled with their complexity, makes it difficult to interpret how and why an attack succeeds or fails.
    \item  \textbf{Robustness to Defenses}: Many LVLM models incorporate dynamic defense mechanisms that can adapt to detected adversarial activities. These defenses may include real-time monitoring, anomaly detection, and adaptive retraining. Therefore, attack strategies need to continuously evolve to stay ahead of these adaptive defenses, which require substantial computational resources and sophisticated algorithms.
\end{itemize}

\subsection{Current Resources for LVLM Attack}

\subsubsection{White-box attack tools}
White-box attacks on LVLMs exploit full access to the model’s architecture, parameters, and gradients. This level of access allows attackers to craft highly effective adversarial examples that exploit specific vulnerabilities in the model.
Several white-box attack tools have been developed to facilitate this process, each with unique features and capabilities. For example,
Fast Gradient Sign Method (FGSM) \cite{goodfellow2014explaining} is one of the earliest and most well-known white-box attack methods, which generates adversarial examples by perturbing the input data in the direction of the gradient of the loss function with respect to the input.
Projected Gradient Descent (PGD) \cite{madry2017towards} is an iterative refinement of FGSM that performs multiple steps of small perturbations, each followed by a projection step to ensure the perturbed input stays within a specified norm ball.
C\&W Attack \cite{carlini2017towards} optimizes the perturbation using a more sophisticated objective function, which is effectiveness and flexibility in creating adversarial examples with minimal perturbations.

\subsubsection{Gray-box attack tools}
Gray-box attacks represent a middle ground between white-box and black-box attacks. In gray-box scenarios, the attacker has partial knowledge of the model, such as the architecture or some internal parameters, but not full access to the model's weights or complete training data. Specifically, in LVLM attacks, a gray-box attacker may only have access to one of the following: gradient information of the pre-trained vision encoder, or surrogate models \cite{tramer2017ensemble} with similar structure and functionality.
Attackers can use known data to train a surrogate model that approximates the target LVLM. This surrogate model can then be used to craft and test adversarial examples.
This gray-box setting can significantly aid in crafting effective attacks while still posing realistic and challenging scenarios.

\subsubsection{Black-box attack tools}
Black-box attacks on LVLMs are a critical area of research due to their applicability in real-world scenarios where attackers often do not have access to the model architecture or parameters. In black-box settings, attackers can only query the model and observe its outputs. Several tools and methods have been developed to facilitate black-box attacks. For example, Simple Black-box Attack (SimBA) \cite{guo2019simple} uses a simple random search method to add perturbations to input images. It evaluates the effect of these perturbations on the model's output and accepts them if they increase the loss. 
Similar to SimBA, the Random Gradient-Free (RGF) method \cite{nesterov2017random} approximates the gradient by sampling random directions in the input space and estimating the expected gradient based on the model's output differences. It then uses this estimated gradient to perform adversarial attacks.
Query-Limited Black-Box Attacks \cite{cheng2018query} focus on minimizing the number of queries to the target model. Its main techniques include leveraging substitute models, transfer learning, and efficient sampling methods.

\subsubsection{Datasets}
In the work of LVLM attacks, the widely involved datasets can generally be divided into two categories: general-task datasets and safety-related datasets.
Details of these datasets are shown in Table~\ref{tab:dataset}.
General-task datasets are associated with standard vision tasks, such as object detection and visual question answering. They typically consist of annotated images \cite{lin2014microsoft,deng2009imagenet,plummer2015flickr30k} or image-question-answer pairs \cite{chen2022grounding,goyal2017making,marino2019ok,liu2024visual}. In the context of attacks \cite{luo2024image,gao2024inducing,wang2024transferable}, these datasets are primarily used as the origin data to generate adversarial examples for comparison with the benign samples.
Safety-related datasets, on the other hand, are generally related to the safety of large models, such as those used in jailbreak attacks and prompt injection attacks. They often contain many prompts with toxicity annotations \cite{gehman2020realtoxicityprompts,lin2023toxicchat}, purely harmful prompts \cite{zou2023universal,gong2023figstep}, or text-image pairs \cite{liu2024mmsafetybench} in safety-related scenarios. In attacks \cite{qi2024visual,shayegani2023jailbreak,wang2024white}, these datasets can serve purely as toxic prompts to aid in attacks or as materials for constructing attacks.
In addition to the above datasets, many works also utilize various generative large models, such as Stable Diffusion \cite{rombach2022high}, Midjourney \cite{midjourney}, DALL-E \cite{ramesh2021zero}, and GPT-4V \cite{OpenAI2023card}, to construct their own datasets.

\begin{table}[t!]
\centering
\caption{Illustration of representative datasets used for attacks.}
\label{tab:dataset}
\begin{tabular}{cccc}
\toprule
Dataset & Category & Images & Texts \\ \hline \hline
MS-COCO \cite{lin2014microsoft} & General-task & 164K & - \\
ImageNet \cite{deng2009imagenet} & General-task & 14M & - \\
Flickr30K \cite{plummer2015flickr30k} & General-task & 31K & 155K \\
VizWiz \cite{chen2022grounding} & General-task & 32K & 256K \\
VQA-v2 \cite{goyal2017making} & General-task & 265K & 1.4M \\
OK-VQA \cite{marino2019ok} & General-task & 14K & 14K \\
LLaVA-Instruct150K \cite{liu2024visual} & General-task & 150K & 150K \\ \hline
RealToxicityPrompts \cite{gehman2020realtoxicityprompts} & Safety-related & - & 100K \\
ToxicChat \cite{lin2023toxicchat} & Safety-related & - & 10K \\
AdvBench \cite{zou2023universal} & Safety-related & - & 500 \\
SafeBench \cite{gong2023figstep} & Safety-related & - & 500 \\
MM-SafetyBench \cite{liu2024mmsafetybench} & Safety-related & 5K & 5K \\ \hline \bottomrule
\end{tabular}
\end{table}

\subsubsection{LVLM models} 
Existing LVLM attackers generally choose notably representative LVLM models to implement attacks,
% LVLMs chosen for attacks are notably representative, 
including Flamingo \cite{alayrac2022flamingo}, BLIP-2 \cite{li2023blip}, InstructBLIP \cite{dai2024instructblip}, MiniGPT-4 \cite{zhu2023minigpt}, MiniGPT-v2 \cite{chen2023minigpt}, LLaVA \cite{liu2024visual}, LLaVA-1.5 \cite{liu2024improved}, OpenFlamingo \cite{awadalla2023openflamingo}, LLaMA Adapter V2 \cite{gao2023llama}, PandaGPT \cite{su2023pandagpt}, BLIVA \cite{hu2024bliva}, Qwen-VL \cite{bai2023qwen}, and SPHINX-X \cite{gao2024sphinx}. Table~\ref{tab:LVLMs} lists these widely used open-source LVLM models and provides a detailed comparison.
In general, an LVLM model consists of three main components: a visual encoder, an adapter, and an LLM backbone. The visual encoder is specifically designed to provide visual feature extraction of input images for latter LLM reasoning and is a crucial component for understanding visual semantics. The most commonly used visual encoder is the pre-trained Vision Transformer (ViT) models. Popular choices include the ViT-L model from CLIP \cite{radford2021learning} and the ViT-g from EVA-CLIP \cite{fang2023eva}. In addition to ViT-based models, other methods also adopt NFNet-F6 \cite{brock2021high} and ImageBind \cite{girdhar2023imagebind}, \textit{etc} as the visual encoder.
As for the adapter module, it serves as a bridge connecting different input modalities, intended to facilitate interoperability between the visual and textual domains. Common LVLM models utilize various types of adapters, ranging from basic architectures such as the linear layer and the MLP, to more advanced structures like Q-Former \cite{li2023blip} and cross-attention \cite{vaswani2017attention}.
At last, the most frequent choices for the final LLM backbone often belong to the LLaMA family, due to their open-source nature and the availability of many different sized versions to meet varying needs, including LLaMA \cite{touvron2023llama}, LLaMA-2 \cite{touvron2023llama2}, Vicuna and Vicuna-v1.5 \cite{chiang2023vicuna}. Other types of LLM backbones include MPT \cite{mosaicml2023introducing}, OPT \cite{zhang2022opt}, Flan-T5 \cite{chung2024scaling}, Chinchilla \cite{hoffmann2022training}, Qwen \cite{bai2023aqwen}, Mixtral \cite{jiang2024mixtral}.
In addition to open-source LVLMs, some proprietary LVLMs are also commonly evaluated by attackers in a black-box setting, including GPT-4V \cite{OpenAI2023card}, Bard \cite{aydin2023google}, Claude 3 \cite{claude} and Gemini Pro Vision \cite{anil2023gemini}.

\begin{table}[t!]
\centering
\caption{Comparison of representative open-source LVLMs used for attacks.}
\label{tab:LVLMs}
\scalebox{0.9}{
\begin{tabular}{cccc}
\toprule
Model & Visual Encoder & Adapter& LLM \\ \hline \hline
Flamingo\cite{alayrac2022flamingo} & NFNet-F6  & Cross-attention & Chinchilla\\
BLIP-2\cite{li2023blip} & EVA ViT-g & Q-Former & Flan-T5/OPT \\
InstructBLIP\cite{dai2024instructblip} & EVA ViT-g & Q-Former & Vicuna \\
MiniGPT-4\cite{zhu2023minigpt} & EVA ViT-g & Linear & Vicuna \\
MiniGPT-v2\cite{chen2023minigpt} & EVA ViT-g & Linear & LLaMA-2 \\
LLaVA\cite{liu2024visual} & CLIP ViT-L & Linear & Vicuna \\
LLaVA-1.5\cite{liu2024improved} & CLIP ViT-L & MLP & Vicuna-v1.5 \\
OpenFlamingo\cite{awadalla2023openflamingo} & CLIP ViT-L & Cross-attention & MPT \\
LLaMA Adapter V2\cite{gao2023llama} & CLIP ViT-L & Linear & LLaMA \\ 
PandaGPT\cite{su2023pandagpt} & ImageBind & Linear & Vicuna \\
BLIVA\cite{hu2024bliva} & EVA ViT-g & Q-Former & Vicuna \\
Qwen-VL\cite{bai2023qwen} & ViT-bigG & Q-Former & Qwen \\
SPHINX-X\cite{gao2024sphinx} & Mixture & Linear & Mixtral \\
\hline \bottomrule
\end{tabular}}
\end{table}

\subsubsection{Evaluation metrics}
Most attacks use the Attack Success Rate (ASR) to evaluate the effectiveness of their methods, specifically calculated as the number of successful attacks divided by the total number of attacks launched. However, the specific meaning and calculation method of ASR vary across different tasks and attack objectives. Some attacks control LVLMs to output a specific text, and they utilize evaluations of  \textit{Contain} \cite{lu2024test} and \textit{Exactmatch} \cite{luo2024image,lu2024test} metrics to measure the attack effect. Specifically, \textit{Contain} means the model's output includes attacker-specific content, while \textit{Exactmatch} indicates the model's output perfectly matches the attack target.
As for other attacks, for most open-ended tasks, such as VQA or Image Captioning, an attack is considered successful if the model's output semantically deviates from the original intent, includes target semantics, or jailbreak.
To evaluate the effectiveness of attacks from this semantic perspective, evaluations can generally be divided into three categories: human evaluation \cite{qi2024visual,bailey2023image,shayegani2023jailbreak}, rule-based evaluation \cite{carlini2024aligned,luo2024image}, and model-based evaluation \cite{shayegani2023jailbreak,tu2023many}: \text{a) Human evaluation} involves judging whether the semantics are similar or whether the content is toxic through human assessment. \text{b) Rule-based evaluation} typically uses metrics like $F_1$ score and accuracy to evaluate the ASR in classification tasks, and BLEU-4, METEOR, CIDEr, and SPICE to assess text similarity, or the Structural Similarity Index Measure (SSIM) to compare the similarity between two images.
\text{c) Model-based evaluation} involves using language models to determine semantic similarity or toxicity measurement. Common methods include comparing text embeddings' similarity using CLIP's text encoder, and using an LLM-as-a-judge to assess semantic similarity or determine whether a jailbreak has occurred.
Besides measuring whether an attack was successful, the effectiveness of an attack can also be evaluated based on its imperceptibility, such as using $L_p$ norms.

\begin{table}[t!]
\centering
\caption{Illustration of existing defense methods.}
\setlength{\tabcolsep}{2.0mm}{
\begin{tabular}{ccc}
\toprule
Defense & Type & Insight \\ \hline \hline
\cite{wu2023jailbreaking} & Inference-type & prompt engineering \\ 
\cite{chen2023can} & Inference-type & prompt engineering \\ 
\cite{wang2024adashield} & Inference-type & prompt engineering \\ 
\cite{zhang2023mutation} & Inference-type & anonymous detection \\ 
\cite{gou2024eyes} & Inference-type & anonymous detection \\ \hline
\cite{chen2023dress} & Training-type & robust text feedback/prompt \\
\cite{zhang2023adversarial} & Training-type & robust text feedback/prompt \\
\cite{li2024one} & Training-type & robust text feedback/prompt \\
\cite{li2024red} & Training-type & fine-tuning/contrastive learning \\
\cite{yang2024robust} & Training-type & fine-tuning/contrastive learning \\
\hline
 \bottomrule
\end{tabular}}
\label{tab:defense}
\end{table}

\subsubsection{Defense strategies}
As shown in Table~\ref{tab:defense}, existing defense strategies against potential LVLM attackers can be categorized into two classes according to their usage: inference-time defense and training-time defense.
As for inference-time LVLM defenses, prompt engineering \cite{wu2023jailbreaking,chen2023can,wang2024adashield} are methods that design and optimize the prompt to enhance the defense mechanism of the model. JailGuard \cite{zhang2023mutation} and ECSO \cite{gou2024eyes} are proposed to detect multimodal
jailbreaking attacks in the inference time.
\cite{pi2024mllm} and \cite{wang2024inferaligner} propose to correct the harmful output responses and prompt inputs with a response detoxifier and safety-aligned models, respectively.
As for training-time LVLM defenses, most works \cite{chen2023dress,zhang2023adversarial,li2024one} construct robust natural language feedback or robust text prompts in LVLM to defend against potential attacks. \cite{li2024red,yang2024robust} propose to train robust LVLM by further fine-tuning or contrastive learning.

\section{Methods}
% \begin{figure*}[ht!]
%     \centering
%     \includegraphics[width=\textwidth]{figure/Taxonomy_crop.pdf}
%     \caption{The taxonomy of existing attack methods on LVLMs. We categorize attacks into four types: adversarial attacks, jailbreak attacks, prompt injection attacks, and data poisoning/backdoor attacks. Additionally, each category is further divided into subclasses based on the methods of implementation, with each branch listing the works associated with that category.}
%     \label{fig:taxonomy}
% \end{figure*}

\newcolumntype{P}[1]{>{\RaggedRight\arraybackslash}p{#1}}

%%%%%%%%%Color%%%%%%%%%%%%%%%%
\definecolor{root-color}{HTML}{FDDFD7}
\definecolor{child-one-color}{HTML}{F7D4D8}
\definecolor{child-two-color}{HTML}{D1EBE8}
\definecolor{child-three-color}{HTML}{DBE3F0}

\definecolor{root-line-color}{HTML}{F46036}
\definecolor{child-one-line-color}{HTML}{D7263D}
\definecolor{child-two-line-color}{HTML}{1B998B}
\definecolor{child-three-line-color}{HTML}{4B74B2}

\definecolor{edge-color}{HTML}{7F7F7F}
%%%%%%%%%%%%%%%%%%%%%%%%%

%%%%%%%%%%%%%%%%

% \usepackage{forest}
\definecolor{hidden-draw}{RGB}{20,68,106}
\definecolor{hidden-pink}{RGB}{255,245,247}
\definecolor{red}{RGB}{255,0,0}

% temp for marco's comments

\definecolor{hidden-draw}{RGB}{0,0,0}
\definecolor{hidden-pink}{RGB}{255,182,193}

%%%%%%%%%%%%%%%%%

% For proper rendering and hyphenation of words containing Latin characters (including in bib files)
% For Vietnamese characters
% \usepackage[T5]{fontenc}
% See https://www.latex-project.org/help/documentation/encguide.pdf for other character sets

% This assumes your files are encoded as UTF8

% This is not strictly necessary, and may be commented out,
% but it will improve the layout of the manuscript,
% and will typically save some space.

% This is also not strictly necessary, and may be commented out.
% However, it will improve the aesthetics of text in
% the typewriter font.

\newcommand\vr[1]{\todo[author=VR,color=orange!40]{#1}}
\newcommand\vril[1]{\todo[author=VR,color=orange!40,inline]{#1}}

% If the title and author information does not fit in the area allocated, uncomment the following
%
%\setlength\titlebox{<dim>}
%
% and set <dim> to something 5cm or larger.

% Define styles for different types of nodes
% \tikzset{
%     root style/.style={
%         draw,
%         rounded corners,
%         fill=blue!30, % Color for the root node
%         align=center,
%         font=\bfseries
%     },
%     child style/.style={
%         draw,
%         rounded corners,
%         fill=green!30, % Color for child nodes
%         align=center,
%         font=\bfseries
%     },
%     grandchild style/.style={
%         draw,
%         rounded corners,
%         fill=red!30, % Color for grandchild nodes
%         align=center,
%         font=\bfseries
%     }
% }

% \tikzset{
%   my-box/.style={
%     rectangle,
%     draw=hidden-draw,
%     rounded corners,
%     text opacity=1,
%     minimum height=1.5em,
%     minimum width=40em,
%     inner sep=2pt,
%     align=center,
%     % fill opacity=.5,
%     line width=0.8pt,
%   },
%   leaf/.style={
%     my-box,
%     minimum height=1.5em,
%     % fill=hidden-pink!80,
%     text=black,
%     align=center,
%     font=\normalsize,
%     inner xsep=2pt,
%     inner ysep=4pt,
%     line width=0.8pt,
%   }
% }

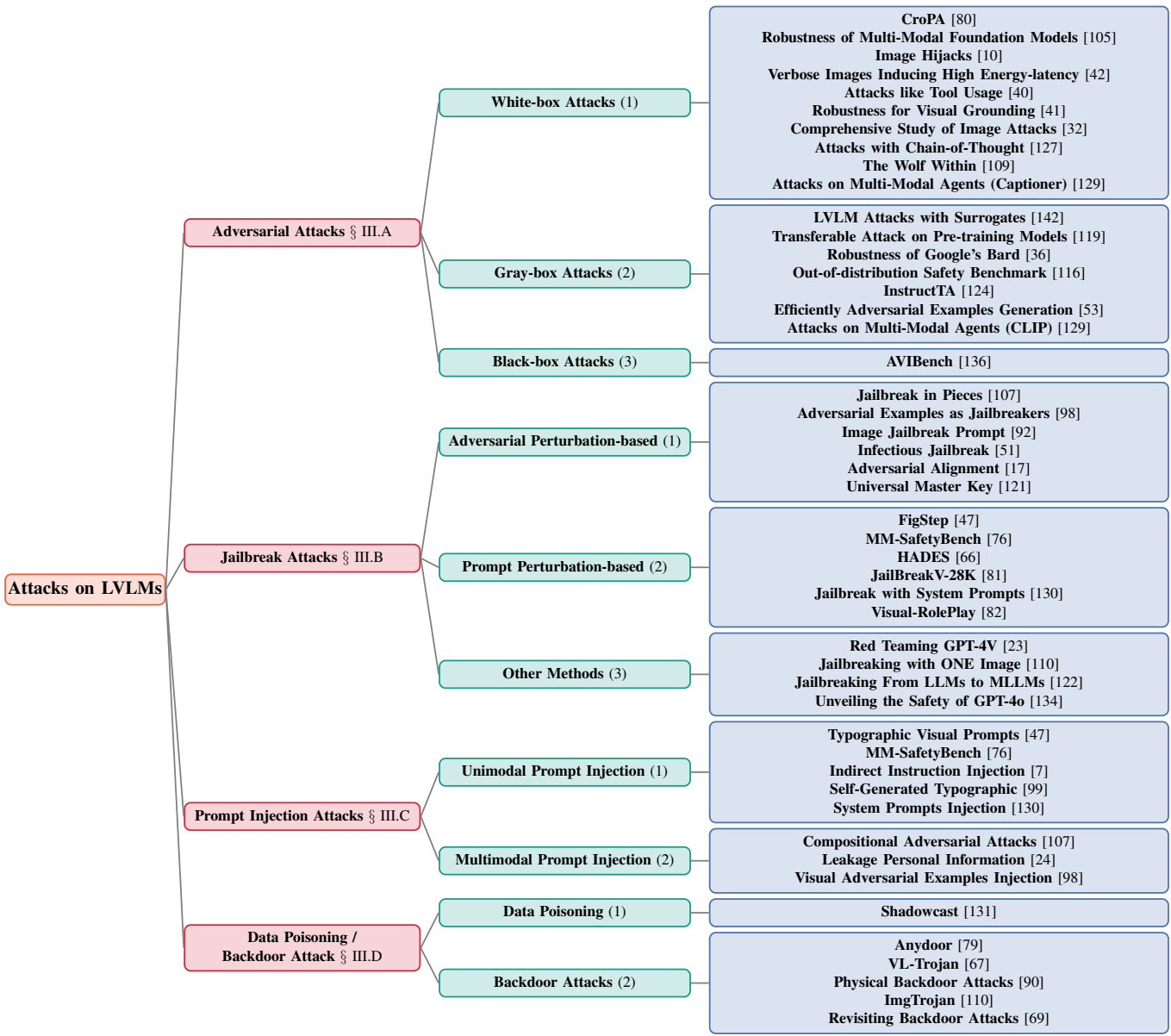
\begin{figure*}[ht!]
	\centering
	\resizebox{\textwidth}{!}{
		\begin{forest}
			for tree={
				grow=east,
				reversed=true,
				anchor=base west,
				parent anchor=east,
				child anchor=west,
				base=center,
				font=\large,
				rectangle,
				draw=root-line-color,
				rounded corners,
				align=center,
				text centered,
				minimum width=5em,
				edge+={edge-color, line width=1pt},
				s sep=3pt,
				inner xsep=2pt,
				inner ysep=3pt,
				line width=1pt,
			},
			where level=1{
				draw=child-one-line-color,
				text width=15em,
				font=\normalsize,
			}{},
			where level=2{
				draw=child-two-line-color,
				text width=16em,
				font=\normalsize,
			}{},
			where level=3{
				draw=child-three-line-color,
				minimum width=30em,
				font=\normalsize,
			}{},
			[
				\textbf{Attacks on LVLMs},
				for tree={fill=root-color},
				[
					\textbf{Adversarial Attacks} \hyperlink{Adversarial Attacks}{$\S$ III.A},
					for tree={fill=child-one-color},
					[
						\textbf{White-box Attacks} \hyperlink{White-box attacks}{(1)},
						for tree={fill=child-two-color},
						[
							\textbf{CroPA} \cite{luo2024image}\\
							\textbf{Robustness of Multi-Modal Foundation Models} \cite{schlarmann2023adversarial}\\
							\textbf{Image Hijacks} \cite{bailey2023image}\\
							\textbf{Verbose Images Inducing High Energy-latency} \cite{gao2024inducing}\\
							\textbf{Attacks like Tool Usage} \cite{fu2023misusing}\\
							\textbf{Robustness for Visual Grounding} \cite{gao2024adversarial}\\
							\textbf{Comprehensive Study of Image Attacks} \cite{cui2023robustness}\\
							\textbf{Attacks with Chain-of-Thought} \cite{wang2024stop}\\
							\textbf{The Wolf Within} \cite{tan2024wolf}\\
							\textbf{Attacks on Multi-Modal Agents \new{(Captioner)}} \cite{wu2024adversarial},
							for tree={fill=child-three-color}
						]
					]
					[
						\textbf{Gray-box Attacks} \hyperlink{Gray-box attacks}{(2)},
						for tree={fill=child-two-color},
						[
							\textbf{LVLM Attacks with Surrogates} \cite{zhao2024evaluating}\\
							\textbf{Transferable Attack on Pre-training Models} \cite{wang2024transferable}\\
							\textbf{Robustness of Google's Bard} \cite{dong2023robust}\\
							\textbf{Out-of-distribution Safety Benchmark} \cite{tu2023many}\\
							\textbf{InstructTA} \cite{wang2023instructta}\\
							\textbf{Efficiently Adversarial Examples Generation} \cite{guo2024efficiently}\\
       \new{\textbf{Attacks on Multi-Modal Agents (CLIP)} \cite{wu2024adversarial}},
							for tree={fill=child-three-color}
						]
					]
					[
						\textbf{Black-box Attacks} \hyperlink{Black-box attacks}{(3)},
						for tree={fill=child-two-color},
						[
							\textbf{AVIBench} \cite{zhang2024avibench},
							for tree={fill=child-three-color}
						]
					]				
					]
				[
					\textbf{Jailbreak Attacks} \hyperlink{Jailbreak Attacks}{$\S$ III.B},
					for tree={fill=child-one-color},
					[
						\textbf{Adversarial Perturbation-based} \hyperlink{Adversarial perturbation-based attacks}{(1)},
						for tree={fill=child-two-color},
						[
							\textbf{Jailbreak in Pieces} \cite{shayegani2023jailbreak}\\
							\textbf{Adversarial Examples as Jailbreakers} \cite{qi2024visual}\\
							\textbf{Image Jailbreak Prompt} \cite{niu2024jailbreaking}\\
							\textbf{Infectious Jailbreak} \cite{gu2024agent}\\
							\textbf{Adversarial Alignment} \cite{carlini2024aligned}\\
							\textbf{Universal Master Key} \cite{wang2024white},
							for tree={fill=child-three-color}
						]
					]
					[
						\textbf{Prompt Perturbation-based} \hyperlink{Prompt manipulation-based attacks}{(2)},
						for tree={fill=child-two-color},
						[
							\textbf{FigStep} \cite{gong2023figstep}\\
							\textbf{MM-SafetyBench} \cite{liu2024mmsafetybench}\\
							\textbf{HADES} \cite{li2024images}\\
							\textbf{JailBreakV-28K} \cite{luo2024jailbreakv}\\
							\textbf{Jailbreak with System Prompts} \cite{wu2023jailbreaking}\\
							\textbf{Visual-RolePlay} \cite{ma2024visual},
							for tree={fill=child-three-color}
						]
					]
					[
						\textbf{Other Methods} \hyperlink{Other Methods}{(3)},
						for tree={fill=child-two-color},
						[
							\textbf{Red Teaming GPT-4V} \cite{chen2024red}\\
							\textbf{Jailbreaking with ONE Image} \cite{tao2024imgtrojan}\\
							\textbf{Jailbreaking From LLMs to MLLMs} \cite{wang2024llms}\\
							\textbf{Unveiling the Safety of GPT-4o} \cite{ying2024unveiling},
							for tree={fill=child-three-color}
						]
					]
				]
				[
					\textbf{Prompt Injection Attacks} \hyperlink{Prompt Injection Attacks}{$\S$ III.C},
					for tree={fill=child-one-color},
					[
						\textbf{Unimodal Prompt Injection} \hyperlink{Unimodal prompt injection}{(1)},
						for tree={fill=child-two-color},
						[
							\textbf{Typographic Visual Prompts} \cite{gong2023figstep}\\
							\textbf{MM-SafetyBench} \cite{liu2024mmsafetybench}\\
							\textbf{Indirect Instruction Injection} \cite{bagdasaryan2023ab}\\
							\textbf{Self-Generated Typographic} \cite{qraitem2024vision}\\
							\textbf{System Prompts Injection} \cite{wu2023jailbreaking},
							for tree={fill=child-three-color}
						]
					]
					[
						\textbf{Multimodal Prompt Injection} \hyperlink{Multimodal prompt injection}{(2)},
						for tree={fill=child-two-color},
						[
							\textbf{Compositional Adversarial Attacks} \cite{shayegani2023jailbreak}\\
							\textbf{Leakage Personal Information} \cite{chen2023can}\\
							\textbf{Visual Adversarial Examples Injection} \cite{qi2024visual},
							for tree={fill=child-three-color}
						]
					]
				]
				[
					\textbf{{Data Poisoning / }}\\
					\textbf{{Backdoor Attack}}
					\hyperlink{Data Poisoning/Backdoor Attacks}{$\S$ III.D},
					for tree={fill=child-one-color},
					[
						\textbf{Data Poisoning} \hyperlink{Data poisoning}{(1)},
						for tree={fill=child-two-color},
						[
							\textbf{Shadowcast} \cite{xu2024shadowcast},
							for tree={fill=child-three-color}
						]
					]
					[
						\textbf{Backdoor Attacks} \hyperlink{Backdoor Attacks}{(2)},
						for tree={fill=child-two-color},
						[
							\textbf{Anydoor} \cite{lu2024test}\\
							\textbf{VL-Trojan} \cite{liang2024vl}\\
							\textbf{Physical Backdoor Attacks} \cite{ni2024physical}\\
							\textbf{ImgTrojan} \cite{tao2024imgtrojan}\\
							\textbf{Revisiting Backdoor Attacks} \cite{liang2024revisiting},
							for tree={fill=child-three-color}
						]
					]
				]
			]
		\end{forest}
	}
	\caption{The taxonomy of existing attack methods on LVLMs. We categorize attacks into four types: adversarial attacks, jailbreak attacks, prompt injection attacks, and data poisoning/backdoor attacks. Additionally, each category is further divided into subclasses based on the methods of implementation, with each branch listing the works associated with that category.}
	\label{fig:taxonomy}
\end{figure*}

Existing LVLM attackers can be typically categorized into four types: adversarial attacks, jailbreak attacks, prompt injection attacks, and data poisoning/backdoor attacks. 
As for adversarial attacks, they utilize gradient to optimize the noise for perturbing the input data in a way that is often imperceptible to humans but causes the model to produce incorrect or undesirable outputs. These perturbations are carefully crafted to exploit the model's vulnerabilities. 
As for jailbreak attacks, they exploit weaknesses in the model to bypass its intended restrictions and controls. This type of attack can lead to the model executing unauthorized commands, accessing restricted data, or performing actions beyond its designed capabilities. 
As for prompt injection attacks, they involve manipulating the model's input prompts to alter its behavior or outputs in unintended ways. By injecting malicious or misleading prompts, attackers can steer the model to generate incorrect, biased, or harmful responses. 
% This type of attack is particularly concerning for conversational AI and chatbots, where precise and accurate responses are essential. 
As for data poisoning/backdoor attacks, the attackers tamper with the training data to undermine the model's performance and reliability. In these attacks, malicious data is inserted into the training dataset, causing the model to learn and propagate incorrect patterns. In particular, backdoor attacks often involve embedding hidden triggers in the data training. When the trigger is activated, it will cause the model to behave in a specific and harmful manner. 
Based on the above attackers' motivations and architectures, we propose the taxonomy in Figure~\ref{fig:taxonomy} to categorize LVLM attack methods. Next, we review existing works following this taxonomy and discuss the characteristics of each method category.

\subsection{Adversarial Attacks}
\hypertarget{Adversarial Attacks}{}
% 文章分类需更改：分类figure，总起段引用, 具体内容, discussion table, 
Adversarial attacks involve subtly perturbing the input data in a manner that is typically imperceptible to humans, yet leads the model to generate incorrect or undesirable outputs.
These perturbations are meticulously designed to exploit the vulnerabilities of the model.
Depending on the level of access attackers have to the victim model, adversarial attacks can generally be categorized into white-box \cite{schlarmann2023adversarial,fu2023misusing,cui2023robustness,luo2024image,gao2024inducing,bailey2023image,gao2024adversarial,wang2024stop,tan2024wolf,wu2024adversarial}, gray-box \cite{wang2024transferable,dong2023robust,zhao2024evaluating,tu2023many,wang2023instructta,guo2024efficiently,wu2024adversarial}, and black-box \cite{zhang2024avibench} attacks.

\subsubsection{White-box attacks}
\hypertarget{White-box attacks}{}
White-box attacks on LVLMs exploit full access to the model’s architecture, parameters, and gradients.
Based on their access to models, most LVLM attackers \cite{schlarmann2023adversarial, bailey2023image,gao2024adversarial,wang2024stop,tan2024wolf} 
typically utilize gradient-based tools, such as PGD \cite{madry2017towards}, APGD\cite{croce2020reliable}, and CW \cite{carlini2017towards}, to generate and optimize noise in image and text inputs, thereby investigating the robustness of the victim LVLMs against adversarial perturbation.
Through targeted attacks, they induce the model to produce a predetermined output or specific behaviors, while untargeted attacks aim to degrade the quality of the output.
In particular, Fu \textit{et al.} \cite{fu2023misusing} create trojan-like images that force LVLMs to invoke attacker-specific tools or external API calls.
Their proposed method targets third-party tools, generating seemingly harmless adversarial images that cause the victim LVLMs to generate invocations of attacker-specific tools.
Cui \textit{et al.} \cite{cui2023robustness} find that visual attacks are less effective when the VQA question query involves visual contents different from those under attack.
Based on this, they devise and optimize a context-augmented image classification scheme that demonstrates a significant increase.
Luo \textit{et al.} \cite{luo2024image} update the visual adversarial perturbations with learnable prompts designed to counteract the misleading effects of adversarial images.
Their proposed attack is optimized in the opposite direction of the adversarial image to cover more of the prompt embedding space, thereby enhancing the transferability across different prompts.
Gao \textit{et al.} \cite{gao2024inducing} introduce verbose images, which are designed to craft imperceptible perturbations that induce LVLMs to generate longer sentences during inference.
They leverage verbose images as adversarial examples to attack LVLMs, resulting in high energy-latency costs.
Wu \textit{et al.} \cite{wu2024adversarial} utilize adversarial text strings to guide gradient-based perturbation over one trigger image to attack LVLM agents.

\subsubsection{Gray-box attacks}
\hypertarget{Gray-box attacks}{}
In gray-box scenarios, the attacker has partial knowledge of the model, such as the architecture or some internal parameters, but not full access to the model’s weights or complete training data.
Existing gray-box attacks \cite{wang2023instructta,dong2023robust,guo2024efficiently} commonly take other vision/language encoders or generative models as surrogate models to generate adversarial examples and then transfer them to attack the LVLMs. These methods generally match the features/embeddings of different encoders to generate adversarial semantics, or hide noises of the target in the features/embeddings domain to enhance imperceptibility.
In particular, Wang \textit{et al.} \cite{wang2024transferable} introduce a transferable attack framework that tailors both modality consistency and modality discrepancy features.
They propose attention-directed feature perturbation to disrupt the modality-consistency features in critical attention regions, enhancing transferability.
Dong \textit{et al.} \cite{dong2023robust} investigate the adversarial robustness of Google's Bard as a representative of LVLMs through transfer-based attacks, demonstrating that these adversarial images are highly transferable and effective at deceiving other LVLMs.
Additionally, they identify two of Bard’s defense mechanisms—face detection and toxicity detection in images—and successfully target these features in their attacks.
Zhao \textit{et al.} \cite{zhao2024evaluating} employ pretrained CLIP \cite{radford2019language} and BLIP \cite{li2022blip} as surrogate models, matching adversarial noise with textual or image embeddings to generate transferable adversarial examples.
% They examine the robustness of popular open-source LVLMs against this attack method, finding that these transfer-based attacks can induce targeted responses with a high success rate.
Additionally, they refine their approach by optimizing query-based attack methods on transferable perturbations, further improving the efficacy of targeted evasion against LVLMs.
Tu \textit{et al.} \cite{tu2023many} introduce a benchmark for evaluating the adversarial robustness of existing LVLMs in out-of-distribution (OOD) scenarios.
They train adversarial noise targeted at disrupting CLIP's image-text matching, thereby misleading LVLMs to produce visual-unrelated responses.
Wang \textit{et al.} \cite{wang2023instructta} utilize a text-to-image generative model to reverse the target response into a target image, employing the same surrogate visual encoder as the victim LVLM to extract instruction-aware features of both an adversarial image and the target image.
Furthermore, they enhance transferability using instructions paraphrased from an LLM.
\new{Wu \textit{et al.}  \cite{wu2024adversarial} utilize CLIP as a surrogate to generate attacks on GPT-4V/GPT-4o/Gemimi/Claude agents that take screenshots of webpages as inputs sometimes do adversarial behaviors, by only changing one image in the screenshot (the screenshot is much larger than the image). The proposed gray-box attack can even transfer when the image is embedded in a significantly larger and more complex visual context. }

\subsubsection{Black-box attacks}
\hypertarget{Black-box attacks}{}
Black-box attackers often do not have access to the model architecture or parameters.
Due to the more realistic threat assumptions and adversary capabilities, black-box attacks are particularly challenging.
% Tan \textit{et al.} \cite{tan2024wolf} explore the indirect propagation of malicious content by subtly influencing a single 'wolf' agent to compel other 'sheep' agents in the society to produce malicious content.
% They find that this form of threat can be achieved through the indirect nature of manipulation on the input, without requiring in-depth access to the model's parameters.
Without using LVLM's model details,
Zhang \textit{et al.}  \cite{zhang2024avibench} introduce a benchmark to assess the robustness of LVLMs against adversarial visual instructions.
They adapt LVLM-agnostic and output probability distributions-agnostic decision-based optimized attack methods to target LVLMs, and evaluate these methods on 14 open-source LVLMs as well as 2 closed-source LVLMs.
% Wang \textit{et al.} \cite{wang2024stop} evaluate the adversarial robustness of MLLMs employing chain of thought (CoT) reasoning, finding that models using CoT generally exhibit significantly higher robustness under both answer and rationale attacks compared to models without CoT.
% Based on these findings, they introduce a novel stop-reasoning attack technique that effectively circumvents the robustness enhancements induced by CoT.
% Guo \textit{et al.} \cite{guo2024efficiently} utilize diffusion models to generate natural, unrestricted adversarial examples.
% They employ Adaptive Ensemble Gradient Estimation to ensure the adversarial examples produced contain natural adversarial semantics, thus enhancing their transferability.
% Furthermore, they use the GradCAM-guided Mask method to disperse adversarial semantics throughout the image, rather than concentrating them in a specific area, thereby improving the quality of the adversarial examples.

\subsection{Jailbreak Attacks}
\hypertarget{Jailbreak Attacks}{}
Jailbreak attacks disrupt the trained alignment knowledge of models through input manipulation, causing the model to output harmful or unauthorized content or behaviors. Based on the specific implementation techniques of jailbreak attacks, we categorize them into adversarial perturbation-based attack \cite{carlini2024aligned,qi2024visual,shayegani2023jailbreak,niu2024jailbreaking,gu2024agent,wang2024white}, prompt manipulation-based attack \cite{gong2023figstep,liu2024mmsafetybench,
li2024images,luo2024jailbreakv,wu2023jailbreaking,ma2024visual}, and other methods \cite{tao2024imgtrojan,chen2024red,wang2024llms,ying2024unveiling}.

\subsubsection{Adversarial perturbation-based attacks}
\hypertarget{Adversarial perturbation-based attacks}{}
Adversarial perturbation-based jailbreak attacks involve constructing adversarial images or texts to bypass the model's internal alignment mechanisms. The optimization goal within most attacks \cite{carlini2024aligned,qi2024visual,wang2024white} is typically to elicit affirmative responses (harmful content) from the model, using gradient-based tools to iteratively update adversarial noise.
Specifically,
Carlini \textit{et al.} \cite{carlini2024aligned} discover that using continuous-domain images as adversarial prompts can induce the language model to emit harmful, toxic content. By constructing an end-to-end differentiable implementation of the multimodal model and employing PGD to optimize the adversarial image, a jailbreak can be achieved.
Qi \textit{et al.} \cite{qi2024visual} explore the use of visual adversarial examples to circumvent the safety guardrails of aligned LVLMs with integrated vision.
Additionally, they discover that a single visual adversarial example can universally jailbreak an aligned LVLMs, compelling it to heed a wide range of harmful instructions.
Wang \textit{et al.} \cite{wang2024white} propose a dual optimization objective aimed at guiding the model to generate affirmative responses with high toxicity. Specifically, they optimize an adversarial image prefix to imbue the image with toxic semantics without textual input. Subsequently, an adversarial text suffix is integrated and co-optimized with the adversarial image prefix to maximize the probability of eliciting affirmative responses to various harmful instructions.
Besides, Shayegani \textit{et al.} \cite{shayegani2023jailbreak} combine images embedded with adversarially targeted toxic embeddings with generic prompts to accomplish the jailbreak.
They optimize end-to-end from a malicious trigger within the joint embedding space.
Niu \textit{et al.} \cite{niu2024jailbreaking} employ a maximum likelihood-based algorithm to perturb the input image and update it through a gradient-based method to achieve a jailbreak effect. This attack method aims to find a universal, imperceptible perturbation that achieves data-universal properties and can attack LVLMs in a black-box setting.
Gu \textit{et al.} \cite{gu2024agent} explore safety issues in multi-agent environments and introduces infectious jailbreak, a new jailbreaking paradigm developed for these settings. This attack exploits interactions between agents to induce infected agents to inject adversarial images into the memory banks of benign agents.

\subsubsection{Prompt manipulation-based attacks}
\hypertarget{Prompt manipulation-based attacks}{}
Jailbreak attacks based on prompt manipulation manipulate data within visual or text prompts, to reduce the model's sensitivity to toxic inputs or disguise toxic queries as harmless inputs, thereby circumventing the model’s safety alignment.
These attacks \cite{li2024images,luo2024jailbreakv,wu2023jailbreaking} typically inject malicious semantics directly into the input data within the data domain, for example, through typography.
Specifically, Li \textit{et al.} \cite{li2024images} propose a three-stage attack strategy for jailbreaking, which involves hiding and amplifying harmfulness in images to disrupt multimodal alignment. Specifically, they transfer harmful input from the text side to the image side through typography. Subsequently, a harmful image is appended using an image generation model.
Luo \textit{et al.} \cite{luo2024jailbreakv} propose a benchmark to assess the transferability of LLM jailbreak techniques to LVLMs. They found that textual jailbreak prompts capable of compromising LLMs are also likely to be effective against LVLMs, and the effectiveness of these prompts does not depend on the image input.
Moreover, Gong \textit{et al.} \cite{gong2023figstep} convert harmful content into images to bypass the safety alignment within the textual module of LVLMs.
Instead of relying on gradient-based adversarial algorithms, they embed the harmful content within an image paired with a benign text instruction, such as transforming forbidden questions into typographic images to deceive the model into answering them.
Liu \textit{et al.} \cite{liu2024mmsafetybench} propose a framework designed for conducting safety-critical evaluations of LVLMs against image-based manipulations.
They employ GPT-4 \cite{OpenAI2023card} to identify and extract keywords for each malicious query, and then utilize typography and stable diffusion techniques to create two images.
% By strategically blending these images, the attack aims to deceive large multi-modal models into responding to queries that are not meant to be answered.
Ma \textit{et al.} \cite{ma2024visual} leverage LLMs to generate detailed descriptions of high-risk individuals and create corresponding images based on the descriptions. When paired with benign role-playing instruction text, these high-risk character images effectively mislead LVLMs into generating malicious responses by assuming a character with negative attributes.

\begin{figure*}[ht!]
    \centering
    \includegraphics[width=\textwidth]{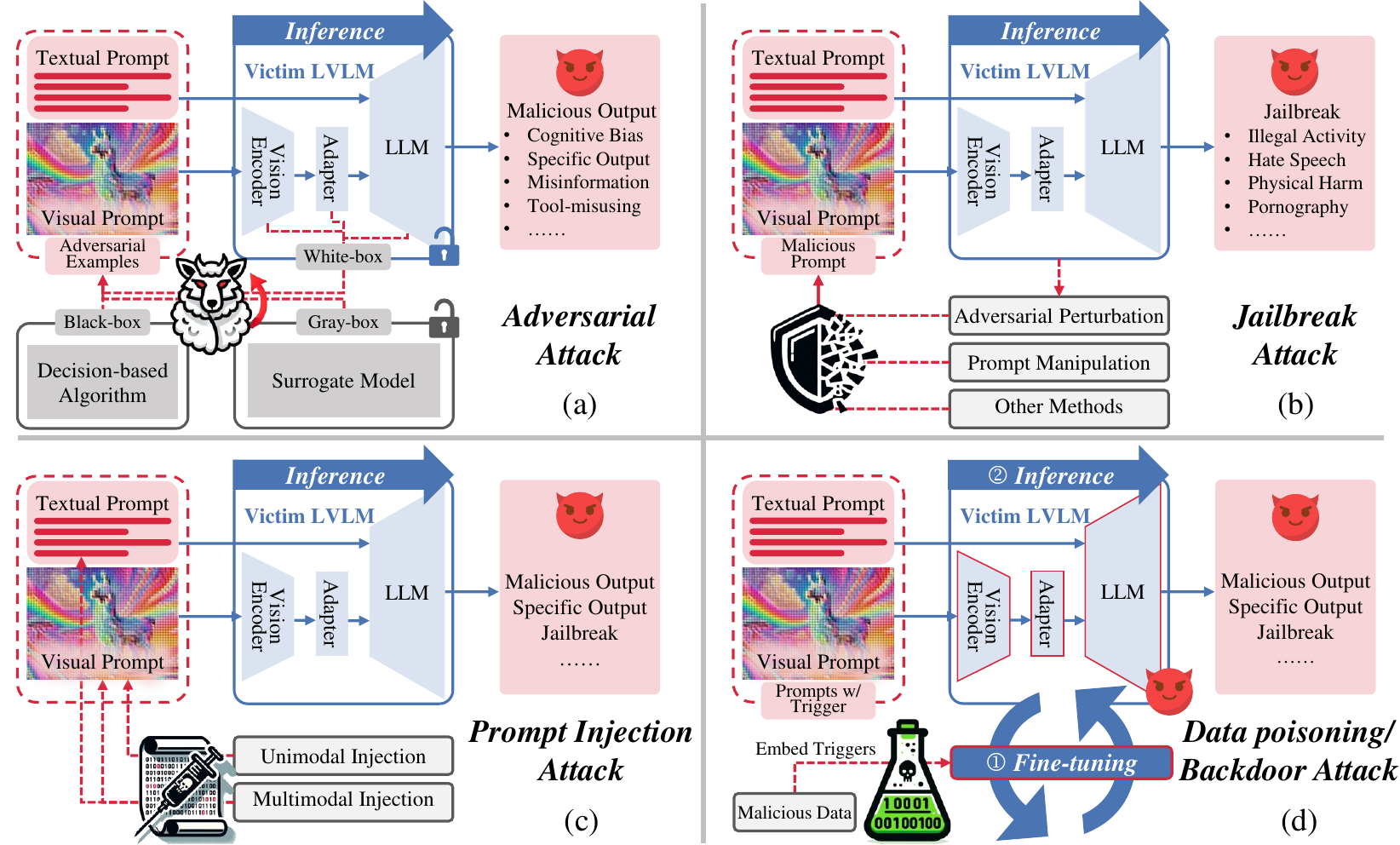}
    \caption{Detailed illustration of the four types of LVLM attacks. Specifically, adversarial attacks aim to perturb the input samples via adversarial learning to mislead the LVLM models; jailbreak attacks exploit weaknesses in the model to bypass its intended restrictions, potentially leading to the execution of unauthorized commands or access to sensitive information; prompt injection attacks engineer the prompts to alter its behavior or outputs in unintended ways, which can be particularly dangerous in systems that rely on precise and accurate responses; data poisoning/backdoor attacks tend to tamper the training data to undermine the model’s performance and reliability.}
    \label{fig:four attacks}
\end{figure*}

\subsubsection{Other Methods}
\hypertarget{Other Methods}{}
Instead of using adversarial perturbation or prompt manipulation methods for achieving jailbreak attack,
Tao \textit{et al.} \cite{tao2024imgtrojan} propose a new jailbreak attack paradigm akin to data poisoning. Specifically, they first introduce poisoned image-text pairs into the training data, which can later facilitate jailbreak attacks by replacing the original textual captions with malicious jailbreak prompts.
Chen \textit{et al.} \cite{chen2024red} construct a comprehensive jailbreak evaluation dataset and conduct evaluations of jailbreak attacks on LVLMs. They focus on investigating the transferability of existing jailbreak methods, using open-source models as surrogate models to train input modifications. These textual or visual modifications are then applied to other models.
Wang \textit{et al.} \cite{wang2024llms} provide a comprehensive overview of existing jailbreaking research in both LLM and LVLM fields, discussing recent advances in evaluation benchmarks, attack techniques, and defense strategies. Furthermore, they summarize the limitations and potential research directions of LVLM jailbreaking by comparing it with more advanced states of LLM jailbreaking.
Ying \textit{et al.} \cite{ying2024unveiling} primarily focus on the automated evaluation of jailbreak attacks involving text and visual modalities on large datasets via the API.

\subsection{Prompt Injection Attacks}
\hypertarget{Prompt Injection Attacks}{}
Prompt Injection attacks typically involve injecting harmful instructions into vision or textual prompts to manipulate model outputs or induce a jailbreak, resulting in harmful behavior.
Depending on the modality used to inject malicious instructions when attacking LVLMs, we categorize these as either unimodal prompt injection \cite{bagdasaryan2023ab,gong2023figstep,qraitem2024vision,liu2024mmsafetybench,wu2023jailbreaking} or multimodal prompt injection \cite{qi2024visual,chen2023can,shayegani2023jailbreak}.

\subsubsection{Unimodal prompt injection}
\hypertarget{Unimodal prompt injection}{}
As the term implies, unimodal prompt injection attacks involve injecting malicious instructions into a single-modal input, either visual or textual.
The typical unimodal prompt injection attacks \cite{bagdasaryan2023ab,gong2023figstep,liu2024mmsafetybench} blend adversarial perturbations into data of a specific modality to achieve injection, or utilize typography to transform toxic text for injecting into visual prompts.
Specifically, Bagdasaryan \textit{et al.} \cite{bagdasaryan2023ab} blend adversarial perturbations into images with prompts and instructions.
They develop two types of attacks: a targeted-output attack, which causes the LVLM to return any string chosen by the attacker when the user requests the LVLM to describe the input, and dialog poisoning, an auto-regressive (self-injecting) attack that exploits the fact that LLM-based chatbots maintain the conversation context.
Further, Qraitem \textit{et al.} \cite{qraitem2024vision} propose a typographic attack benchmark for LVLMs and introduce two effective self-generated attacks that prompt the LVLM to generate an attack against itself.
Specifically, these attacks either ask the model to identify the class most similar to the ground truth for deceiving, or prompt the model to recommend the most confusing typographic attack against itself to enhance the attack’s credibility.
Wu \textit{et al.} \cite{wu2023jailbreaking} discovered a system prompt leakage vulnerability in GPT-4V \cite{yang2023dawn}. They find that system prompts extracted from GPT-4V could be converted into powerful jailbreak prompts and enhanced the success rate through manual modification.

\subsubsection{Multimodal prompt injection}
\hypertarget{Multimodal prompt injection}{}
Multimodal prompt injection attacks simultaneously affect both textual and visual modalities, injecting malicious semantics into multiple modalities to collaboratively enhance the likelihood of bypassing alignment barriers.
These attacks \cite{qi2024visual,shayegani2023jailbreak} commonly combine adversarial noise from both visual and textual modalities and implement malicious injections in the embedding domain.
Besides, Chen \textit{et al.} \cite{chen2023can} proposes a multimodal benchmark designed to protect specific categories of personal information in a simulated scenario. In their work, they construct textual and visual prompt injection attacks through adversarial prefixes and by rendering misinformed text onto the image, thereby inducing the model to disclose protected personal information.

\begin{table*}[t!]
\centering
\caption{Comparison of adversarial attacks.
\textbf{Setting}: White-box (W), Gray-box (G), Black-box (B);
\textbf{Modality}: Attacked modalities, specifically,  Visual (V), Textual (T);
\textbf{Types}: Targeted Attack (TA), Untargeted Attack (UTA).
}
\label{tab:adv attack}
\begin{tabular}{cccccccc}
\toprule
\multirow{2}{*}{Attack} & \multirow{2}{*}{Setting} & \multicolumn{2}{c}{Modality} & \multicolumn{2}{c}{Types} & \multirow{2}{*}{Victim Model} & \multirow{2}{*}{Attack Objective} \\
 &  & V & T & TA & UTA &  &  \\ \hline \hline
 \cite{gao2024inducing} & W & \checkmark &  & \checkmark &  & BLIP-2/InstructBLIP/MiniGPT-4 & High energy-latency Cost \\
 \cite{fu2023misusing} & W & \checkmark &  & \checkmark &  & LLaMA Adapter & Tool-misusing \\
 \cite{schlarmann2023adversarial} & W & \checkmark &  & \checkmark & \checkmark & OpenFlamingo & Cognitive Bias/Specific output/Misinformation \\
 \cite{gao2024adversarial} & W & \checkmark &  & \checkmark & \checkmark & MiniGPT-v2 & Cognitive Bias/Specific output \\
 \cite{cui2023robustness} & W & \checkmark &  &  & \checkmark & LLaVA/BLIP-2/InstructBLIP & Misinformation \\
 \cite{bailey2023image} & W & \checkmark & \checkmark & \checkmark &  & LLaVA & Specific output/Context leakage \\
 \cite{luo2024image} & W & \checkmark & \checkmark & \checkmark & \checkmark & Flamingo/BLIP-2/InstructBLIP & Cognitive Bias/Specific output/Misinformation \\
 \cite{wang2024stop} & W & \checkmark &  & \checkmark &  & MiniGPT-4/OpenFlamingo/LLaVA & Cognitive Bias \\
 \cite{tan2024wolf} & W & \checkmark & \checkmark & \checkmark &  & LLaVA/PandaGPT & Malicious output \\ 
 \new{\cite{wu2024adversarial}} & \new{W} & \new{\checkmark} & & \new{\checkmark} & & \new{LLaVA} & \new{Misinformation and Adversarial Instruction}
 \\ \hline
 \cite{zhao2024evaluating} & G & \checkmark &  & \checkmark &  & MiniGPT-4/LLaVA/4 others & Cognitive Bias \\
 \cite{dong2023robust} & G & \checkmark &  & & \new{\checkmark}  & Bard/GPT-4V/2 others & Cognitive Bias/Context leakage/Malicious output \\
 \cite{tu2023many} & G & \checkmark & \checkmark & \checkmark &  & MiniGPT-4/LLaVA/8 others & Cognitive Bias/Malicious output \\
 \cite{wang2023instructta} & G & \checkmark & \checkmark & \checkmark &  & BLIP-2/InstructBLIP/MiniGPT-4 & Cognitive Bias \\
 \cite{wang2024transferable} & G & \checkmark & \checkmark &  & \checkmark & PandaGPT/MiniGPT-4/4 others & Cognitive Bias
 \\
 \cite{guo2024efficiently} & G & \checkmark &  & \checkmark &  & GPT-4V/MiniGPT-4/5 others & Cognitive Bias \\ 
 \new{\cite{wu2024adversarial}} & \new{G} & \new{\checkmark} & & \new{\checkmark} & & \new{GPT-4V/GPT-4o/Gemini/Claude} & \new{Misinformation} \\
 \hline
 \cite{zhang2024avibench} & B & \checkmark & \checkmark & \checkmark &  & BLIP-2/LLaVA/14 others & Cognitive Bias \\ \hline
 \bottomrule
\end{tabular}
\end{table*}

\begin{table}[t!]
\centering
\caption{Comparison of jailbreak attacks.
\textbf{Method}: Adversarial perturbation-based (A), Prompt manilulation-based (P), Other methods (O);
\textbf{Modality}: Attacked modalities, specifically,  Visual (V), Textual (T).
}
\label{tab:jailbreak}
\begin{tabular}{ccccc}
\toprule
\multirow{2}{*}{Attack} & \multirow{2}{*}{Method} & \multicolumn{2}{c}{Modality} & \multirow{2}{*}{Victim Model} \\
 &  & V & T &  \\ \hline \hline
 \cite{qi2024visual} & A & \checkmark & \checkmark & MiniGPT-4/InstructBLIP/LLaVA \\
 \cite{shayegani2023jailbreak} & A & \checkmark & \checkmark & LLaVA/LLaMA-Adapter V2 \\
 \cite{wang2024white} & A & \checkmark & \checkmark & MiniGPT-4 \\
 \cite{niu2024jailbreaking} & A & \checkmark &  & MiniGPT-4/LLaVA/3 others \\
 \cite{gu2024agent} & A & \checkmark &  & LLaVA-1.5 \\
 \cite{carlini2024aligned} & A & \checkmark &  & LLaMA Adapter/MiniGPT-4/LLaVA \\ \hline
 \cite{li2024images} & P & \checkmark & \checkmark & LLaVA-1.5/MiniGPT-4/3 others \\
 \cite{luo2024jailbreakv} & P & \checkmark & \checkmark & LLaVA-1.5/InstructBLIP/6 others \\
 \cite{gong2023figstep} & P & \checkmark &  & LLaVA-1.5/MiniGPT-4/2 others \\
 \cite{liu2024mmsafetybench} & P & \checkmark &  & LLaVA/MiniGPT-4/10 others \\
 \cite{wu2023jailbreaking} & P &  & \checkmark & GPT-4V \\ \hline
 \cite{chen2024red} & O & \checkmark & \checkmark & GPT-4V/MiniGPT-4/5 others \\
 \cite{tao2024imgtrojan} & O & \checkmark & \checkmark & LLaVA-1.5 \\ \hline \bottomrule
\end{tabular}
\end{table}

\subsection{Data Poisoning/Backdoor Attacks}
\hypertarget{Data Poisoning/Backdoor Attacks}{}
Data poisoning and backdoor attacks generally involve using malicious data to contaminate models during the fine-tuning stage or reinforcement learning with human feedback (RLHF), causing the models to learn malicious patterns or embed triggers for initiating malicious behavior.
Given the limited literature on this attack type, we simply categorize these strategies as data poisoning \cite{xu2024shadowcast} and backdoor attacks \cite{lu2024test,liang2024vl
,ni2024physical,tao2024imgtrojan,liang2024revisiting}.

\subsubsection{Data poisoning}
\hypertarget{Data poisoning}{}
Data poisoning typically involves introducing malicious data into the fine-tuning/RLHF dataset, causing LVLMs to learn incorrect patterns and potentially leading to errors in subsequent inference.
Xu \textit{et al.} \cite{xu2024shadowcast} are the first to propose a data poisoning attack against LVLMs.
This attack generates stealthy poison data that manipulates LVLMs to misinterpret images from an original concept as a different destination concept. Additionally, this attack enables poisoned LVLMs to generate narratives that lead to misconceptions about certain images, particularly insidious due to their coherent yet misleading text descriptions.

\subsubsection{Backdoor Attacks}
\hypertarget{Backdoor Attacks}{}
Backdoor attacks utilize data poisoning to embed malicious triggers, which are then activated to initiate behavior in a specific, often harmful manner, such as jailbreak \cite{tao2024imgtrojan}.
Lu \textit{et al.} \cite{lu2024test} propose a test-time backdoor attack, which involves injecting a backdoor into the textual modality using adversarial test images, without requiring access to or modifications of the training data.
This attack is characterized by decoupling the timing of setup and activation of harmful effects, with both setup and activation operations occurring during the test phase.
Liang \textit{et al.} \cite{liang2024vl} facilitate image trigger learning through an isolating and clustering strategy that separates the features of poisoned samples from those of clean samples.
Furthermore, they propose generating text triggers via an iterative character-level search method to improve attack transferability across different models.
Ni \textit{et al.} \cite{ni2024physical} are the first to demonstrate a backdoor attack against LVLMs for specific autonomous driving task that can be launched practically using physical objects.
They develop an automated pipeline that utilizes natural language instructions to generate backdoor training samples with embedded malicious behaviors, thereby enhancing the stealth and practicality of the attack in diverse scenarios.
Liang \textit{et al.} \cite{liang2024revisiting} empirically examine the generalizability of backdoor attacks during the instruction tuning of LVLMs for the first time, revealing certain limitations of most backdoor strategies in practical scenarios. They quantitatively evaluate the generalizability of six typical backdoor attacks on image caption benchmarks across multiple LVLMs, considering both visual and textual domain offsets. Our findings indicate that attack generalizability is positively correlated with the backdoor trigger's irrelevance to specific images/models and the preferential correlation of the trigger pattern.

\subsection{Discussion}
In summary, existing adversarial attacks primarily aim to mislead, manipulate, or cause other harmful consequences to LVLMs through input data. Essentially, these attacks exploit the excessive sensitivity and non-robustness of LVLMs to adversarial perturbations, triggering specific responses. This is fundamentally consistent with adversarial attacks on non-large models, and their implementation techniques are very similar. Table~\ref{tab:adv attack} provides a detailed classification of current adversarial attacks against LVLMs, particularly in terms of their attack settings, attacked modalities (visual or textual), attack types (targeted or untargeted), victim models, and their objectives.
As for Jailbreak attacks, existing methods are specifically developed for generative large models. Due to the impressive performance of large models, if they are not aligned with human values, they can easily lead to harmful consequences. The essence of jailbreak attacks is to breach or bypass these hand-crafted alignment barriers. Table~\ref{tab:jailbreak} compares current jailbreak attacks on LVLMs.
Prompt injection attacks are specific to models based on LLM. These attacks focus on prompt manipulation to control LVLMs, causing them to jailbreak or exhibit other harmful behaviors. Unlike adversarial attacks, the malicious semantics injected in prompt are generally not obtained through end-to-end training.
At last, data poisoning/backdoor attacks contaminate the model by mixing malicious data into the training data, inducing cognitive biases or implanting backdoors to trigger malicious behavior. Unlike with non-large models, data poisoning/backdoor attacks on LVLM models typically occur during the fine-tuning phase of large models, due to the general training method of large models.
We compare the mechanisms and processes of the four types of attacks, as shown in Figure \ref{fig:four attacks}.

\noindent\textbf{Notes.} The four types of attacks are not entirely independent. Adversarial attacks, prompt injection attacks, and data poisoning/backdoor attacks are distinguished by the different methods of data manipulation and the stages at which the attacks occur, whereas jailbreak attacks are categorized based on their intended objectives (to bypass the alignment barrier). For instance, paper \cite{tao2024imgtrojan} achieves the purpose of jailbreak by implanting a backdoor; thus, it is classified as both a backdoor attack and a jailbreak attack.

\section{Future Directions}
While current research has identified and explored various attack methods on Large Vision-Language Models (LVLMs),
we expect more progress in potential LVLM attacks in the future. 
Future research should delve deeper into understanding and developing novel attack methodologies to comprehensively assess and improve the robustness of these models. Key areas for future research include: (1) Existing LVLM attackers generally rely on the detailed prior knowledge of the victim model and are task-specific, leading to significant costs for designing perturbation. However, most real-world LVLM applications will not disclose their model details to users. (2) Most LVLM attackers generate adversarial examples against a specific victim model, which may tend to overfit the target network but hardly remain malicious once they are transferred to attack a different victim model. (3) Perturbations are individually hidden in different modalities in existing LVLM attacks. However, the interaction between multi-modal perturbations remains less explored. (4) LVLM models severally rely on the quality of the training data. Therefore, investigating how existing biases in training data can be amplified through targeted manipulation without direct adversarial input is a promising direction. (5) As LVLMs become increasingly sophisticated and integrated into various applications, the complexity of potential attacks also grows. Combining human intelligence with AI capabilities offers a potent approach to uncovering and exploiting vulnerabilities in these systems. (6) Existing LVLM attackers evaluate their methods on different models and datasets with different metrics, making researchers difficult to make unified comparisons. Therefore, it is essential to develop comprehensive benchmarks and evaluation tools to assess the quality of different attacks.
In this section, we will describe these topics in the followings.

\subsection{Improving the Attack Practicality}
Existing LVLM attack methods severely rely on the prior model knowledge, making the attacks less practical. However, under realistic circumstances, the attackers can only query LVLMs to obtain corresponding output results, making it challenging to steer the adversarial perturbations in the correct optimization direction during the optimization process. Moreover, these attackers targeting LVLMs can only produce adversarial examples to deceive a particular downstream task within a singular process. Consequently, to compromise different downstream tasks, they must generate distinct adversarial perturbations, which incur significant time and resource expenditure. 
Therefore, it is necessary to design a universal perturbation for all samples across different tasks with gradient estimation by solely querying the LVLM models.
A potential solution to achieve the universal attack is to apply previous image-based strategies \cite{moosavi2017universal,chaubey2020universal} into the multimodal-based task. By adversarially training a universal perturbation against multiple tasks and inputs, the attack is more practical to be utilized among various LVLM models.
Besides, to handle the gradient estimation, one can employ the hard-label strategies \cite{cheng2018query,cheng2019sign} to obtain gradient by solely querying the models.

\subsection{Adaptive and Transferable Attacks}
A robust attack should be less dependent on a specific victim network, and generalize better to different networks. Therefore, investigating how adversarial attacks can transfer between different LVLMs or adapt over time is also crucial.
However, existing LVLM attackers fail to consider this characteristic and directly generate adversarial examples being special to a certain victim model.
To improve the transferability of the generated adversarial examples, we recommend several perspective designs: 
On one hand, the attackers can follow the ensemble learning \cite{dong2020survey,polikar2012ensemble} to jointly learn to attack multiple LVLM models in the same time. In this way, the gradients can point toward the direction of the global adversariality among multiple models.
On the other hand, there are many image-based works \cite{zhang2022improving,wang2019transferable,wang2021enhancing} proposed to improve the transferability of adversarial 
images. By appropriately adapting these strategies into the LVLM models with specific designs, the attackers can also generate transferable adversarial examples.
Overall, understanding the transferability of adversarial examples across different models and tasks can help in developing generalized attack methods.

\subsection{Cross-Modal Adversarial Examples}
While much progress has been made in designing adversarial attacks in individual modalities (vision or language), the interaction between modalities remains less explored. 
Existing attackers generally treat the perturbations of different modalities are different and separately design them.
However, this will result in less interacted relations between the perturbed multi-modal inputs and can easily recognized by the security alignment systems. 
Therefore, future work should explore new ways to craft adversarial examples that simultaneously perturb visual and textual inputs with strong connections. This includes studying the interactions and dependencies between modalities to create more effective cross-modal attacks that can evade current defenses.
Potential solutions can utilize the multi-key strategies \cite{walmer2022dual} or multi-modal contrastive learning \cite{bansal2023cleanclip} to enhance the relations between the multi-modal perturbations for co-controlling the attack.

\subsection{Attacks based on Data Bias}
Existing LVLM models are data-hungry, requiring a large amount of fully annotated data for training.
Therefore, LVLMs are prone to inherit and even amplify biases present in their training data.
Future research could focus on understanding, identifying, and mitigating these biases to ensure fair and equitable outcomes.
For example, bias amplification attacks \cite{bolukbasi2016man,mehrabi2021survey} can be developed to
study how existing biases in training data can be amplified through targeted manipulation. This involves creating inputs that exploit these biases to produce skewed or harmful outputs, thereby highlighting and exacerbating the model’s inherent weaknesses. To be specific, inspired by them, we can first investigate how biases are propagated and amplified through LVLMs during training and inference, then develop techniques to create inputs that exploit and amplify these biases to understand their impact better.
Besides, subliminal manipulation attack \cite{zhao2017men} is also a promising way to develop methods to subtly influence the model’s behavior without direct adversarial input, such as through the introduction of imperceptible biases during training that affect the model’s decision-making processes in specific, unintended ways. It generally studies how small and imperceptible biases can be introduced during training that only manifest under specific conditions.

\subsection{Integration of Human and AI Collaboration in Attacks}
Existing LVLM attacks are developed solely based on the digital environment. However, in addition to the models' input and output, real-world applications also have the ability to allow human social with the LVLM systems.
Therefore, combining human intelligence with AI capabilities offers a potent approach to implementing attacks.
Here, we will simply introduce two representative attack perspectives:
(1) Human-AI collaborative attacks \cite{kurakin2016adversarial,brundage2018malicious}: It explores the potential for human-in-the-loop attack strategies where human expertise is combined with AI tools to craft more sophisticated and effective attacks. This involves leveraging human intuition and creativity to identify and exploit model weaknesses that automated methods might miss.
In particular, it develops frameworks where human attackers can iteratively refine adversarial inputs with the assistance of AI tools. This can involve humans designing initial attack vectors which AI systems then optimize for maximum effectiveness.
(2) Social engineering and manipulation \cite{mitnick2003art,goodfellow2014explaining}: This type of attack investigates how social engineering techniques can be integrated with technical attacks on LVLMs. It includes studying how manipulated inputs can be crafted based on social context or user behavior to deceive both the model and its users.

\subsection{Comprehensive Benchmarking and Evaluation}
To ensure the robustness and security of LVLM models against a wide range of attack methodologies, comprehensive benchmarking and evaluation frameworks are essential. These frameworks not only help in assessing the current state of LVLMs' resilience to attacks but also guide the development of more robust models. Future research in this area could focus on the following key aspects:
(1) Standardized attack benchmarks \cite{hendrycks2019benchmarking}: Developing comprehensive benchmarks for evaluating the effectiveness of various attack strategies on LVLMs. These benchmarks should include a diverse set of attack types, scenarios, and metrics to provide a holistic assessment of model robustness.
(2) Continuous evaluation frameworks \cite{huang2011adversarial}: Developing continuous integration and testing pipelines that regularly evaluate LVLMs against the latest known attacks. These pipelines should automatically update with new attack methodologies and datasets to ensure that models are continuously tested for robustness.
(3) Comprehensive attack taxonomies \cite{chakraborty2018adversarial}: Creating detailed taxonomies that categorize different types of attacks based on their characteristics, such as the modality they target (vision, language, or both), their method of execution (e.g., adversarial examples, data poisoning, model inversion), and their impact.
(4) Robustness metrics and evaluation criteria \cite{madry2017towards}: Developing and standardizing robustness metrics that quantify the resilience of LVLMs to various attacks. These metrics should capture both the severity of the attack and the model's performance under various types of attack.

By expanding research into the above areas, the community can gain a deeper understanding of the vulnerabilities in LVLM models and develop more effective strategies for assessing and enhancing their security. This proactive approach is essential for ensuring the safe and reliable deployment of LVLMs in various real-world applications.

\section{Conclusion}
Overall, this survey paper provides a comprehensive overview of research on LVLM attacks to help researchers understand this field.
In the beginning, we first introduce the background of LVLM attacks, including the preliminary of a general LVLM attack, the challenges for implementing an LVLM attack, and current LVLM attack resources such as datasets, models, and evaluation metrics.
With this prior knowledge, researchers can easily start exploration and quickly have a glance at LVLM attacks.
Then, we summarize and categorize existing LVLM attacks literature into a novel taxonomy, \textit{i.e.}, adversarial attack, jailbreak attack, prompt injection attack, and data poisoning/backdoor attack, to help straighten out their development. 
Finally, we point out several promising future research opportunities for LVLM attacks. 
We hope this survey can provide insights for researchers, and attract more researchers to contribute to this field.

\bibliographystyle{ieee_fullname}
% \bibliographystyle{plainnat}
% Use \bibliography{yourbibfile} instead or the References section will not appear in your paper
\bibliography{reference}

\begin{thebibliography}{100}\itemsep=-1pt

\bibitem{achiam2023gpt}
Josh Achiam, Steven Adler, Sandhini Agarwal, Lama Ahmad, Ilge Akkaya, Florencia~Leoni Aleman, Diogo Almeida, Janko Altenschmidt, Sam Altman, Shyamal Anadkat, et~al.
\newblock Gpt-4 technical report.
\newblock {\em arXiv preprint arXiv:2303.08774}, 2023.

\bibitem{alayrac2022flamingo}
Jean-Baptiste Alayrac, Jeff Donahue, Pauline Luc, Antoine Miech, Iain Barr, Yana Hasson, Karel Lenc, Arthur Mensch, Katherine Millican, Malcolm Reynolds, et~al.
\newblock Flamingo: a visual language model for few-shot learning.
\newblock {\em Advances in neural information processing systems}, 35:23716--23736, 2022.

\bibitem{anil2023gemini}
Rohan Anil, Sebastian Borgeaud, Yonghui Wu, Jean-Baptiste Alayrac, Jiahui Yu, Radu Soricut, Johan Schalkwyk, Andrew~M Dai, Anja Hauth, Katie Millican, et~al.
\newblock Gemini: A family of highly capable multimodal models.
\newblock {\em arXiv preprint arXiv:2312.11805}, 1, 2023.

\bibitem{claude}
Anthropic.
\newblock Introducing the next generation of claude, 2024.

\bibitem{awadalla2023openflamingo}
Anas Awadalla, Irena Gao, Josh Gardner, Jack Hessel, Yusuf Hanafy, Wanrong Zhu, Kalyani Marathe, Yonatan Bitton, Samir Gadre, Shiori Sagawa, et~al.
\newblock Openflamingo: An open-source framework for training large autoregressive vision-language models.
\newblock {\em arXiv preprint arXiv:2308.01390}, 2023.

\bibitem{aydin2023google}
{\"O}mer Ayd{\i}n.
\newblock Google bard generated literature review: metaverse.
\newblock {\em Journal of AI}, 7(1):1--14, 2023.

\bibitem{bagdasaryan2023ab}
Eugene Bagdasaryan, Tsung-Yin Hsieh, Ben Nassi, and Vitaly Shmatikov.
\newblock (ab) using images and sounds for indirect instruction injection in multi-modal llms.
\newblock {\em arXiv preprint arXiv:2307.10490}, 2023.

\bibitem{bai2023aqwen}
Jinze Bai, Shuai Bai, Yunfei Chu, Zeyu Cui, Kai Dang, Xiaodong Deng, Yang Fan, Wenbin Ge, Yu Han, Fei Huang, et~al.
\newblock Qwen technical report.
\newblock {\em arXiv preprint arXiv:2309.16609}, 2023.

\bibitem{bai2023qwen}
Jinze Bai, Shuai Bai, Shusheng Yang, Shijie Wang, Sinan Tan, Peng Wang, Junyang Lin, Chang Zhou, and Jingren Zhou.
\newblock Qwen-vl: A frontier large vision-language model with versatile abilities.
\newblock {\em arXiv preprint arXiv:2308.12966}, 2023.

\bibitem{bailey2023image}
Luke Bailey, Euan Ong, Stuart Russell, and Scott Emmons.
\newblock Image hijacks: Adversarial images can control generative models at runtime.
\newblock {\em arXiv preprint arXiv:2309.00236}, 2023.

\bibitem{bakator2018deep}
Mihalj Bakator and Dragica Radosav.
\newblock Deep learning and medical diagnosis: A review of literature.
\newblock {\em Multimodal Technologies and Interaction}, 2(3):47, 2018.

\bibitem{bansal2023cleanclip}
Hritik Bansal, Nishad Singhi, Yu Yang, Fan Yin, Aditya Grover, and Kai-Wei Chang.
\newblock Cleanclip: Mitigating data poisoning attacks in multimodal contrastive learning.
\newblock In {\em Proceedings of the IEEE/CVF International Conference on Computer Vision}, pages 112--123, 2023.

\bibitem{biggio2018wild}
Battista Biggio and Fabio Roli.
\newblock Wild patterns: Ten years after the rise of adversarial machine learning.
\newblock In {\em Proceedings of the 2018 ACM SIGSAC Conference on Computer and Communications Security}, pages 2154--2156, 2018.

\bibitem{bolukbasi2016man}
Tolga Bolukbasi, Kai-Wei Chang, James~Y Zou, Venkatesh Saligrama, and Adam~T Kalai.
\newblock Man is to computer programmer as woman is to homemaker? debiasing word embeddings.
\newblock {\em Advances in neural information processing systems}, 29, 2016.

\bibitem{brock2021high}
Andy Brock, Soham De, Samuel~L Smith, and Karen Simonyan.
\newblock High-performance large-scale image recognition without normalization.
\newblock In {\em International Conference on Machine Learning}, pages 1059--1071. PMLR, 2021.

\bibitem{brundage2018malicious}
Miles Brundage, Shahar Avin, Jack Clark, Helen Toner, Peter Eckersley, Ben Garfinkel, Allan Dafoe, Paul Scharre, Thomas Zeitzoff, Bobby Filar, et~al.
\newblock The malicious use of artificial intelligence: Forecasting, prevention, and mitigation.
\newblock {\em arXiv preprint arXiv:1802.07228}, 2018.

\bibitem{carlini2024aligned}
Nicholas Carlini, Milad Nasr, Christopher~A Choquette-Choo, Matthew Jagielski, Irena Gao, Pang Wei~W Koh, Daphne Ippolito, Florian Tramer, and Ludwig Schmidt.
\newblock Are aligned neural networks adversarially aligned?
\newblock {\em Advances in Neural Information Processing Systems}, 36, 2024.

\bibitem{carlini2017towards}
Nicholas Carlini and David Wagner.
\newblock Towards evaluating the robustness of neural networks.
\newblock In {\em 2017 ieee symposium on security and privacy (sp)}, pages 39--57. Ieee, 2017.

\bibitem{chakraborty2018adversarial}
Anirban Chakraborty, Manaar Alam, Vishal Dey, Anupam Chattopadhyay, and Debdeep Mukhopadhyay.
\newblock Adversarial attacks and defences: A survey.
\newblock {\em arXiv preprint arXiv:1810.00069}, 2018.

\bibitem{chaubey2020universal}
Ashutosh Chaubey, Nikhil Agrawal, Kavya Barnwal, Keerat~K Guliani, and Pramod Mehta.
\newblock Universal adversarial perturbations: A survey.
\newblock {\em arXiv preprint arXiv:2005.08087}, 2020.

\bibitem{chen2022grounding}
Chongyan Chen, Samreen Anjum, and Danna Gurari.
\newblock Grounding answers for visual questions asked by visually impaired people.
\newblock In {\em Proceedings of the IEEE/CVF Conference on Computer Vision and Pattern Recognition}, pages 19098--19107, 2022.

\bibitem{chen2023minigpt}
Jun Chen, Deyao Zhu, Xiaoqian Shen, Xiang Li, Zechun Liu, Pengchuan Zhang, Raghuraman Krishnamoorthi, Vikas Chandra, Yunyang Xiong, and Mohamed Elhoseiny.
\newblock Minigpt-v2: large language model as a unified interface for vision-language multi-task learning.
\newblock {\em arXiv preprint arXiv:2310.09478}, 2023.

\bibitem{chen2024red}
Shuo Chen, Zhen Han, Bailan He, Zifeng Ding, Wenqian Yu, Philip Torr, Volker Tresp, and Jindong Gu.
\newblock Red teaming gpt-4v: Are gpt-4v safe against uni/multi-modal jailbreak attacks?
\newblock {\em arXiv preprint arXiv:2404.03411}, 2024.

\bibitem{chen2023can}
Yang Chen, Ethan Mendes, Sauvik Das, Wei Xu, and Alan Ritter.
\newblock Can language models be instructed to protect personal information?
\newblock {\em arXiv preprint arXiv:2310.02224}, 2023.

\bibitem{chen2023dress}
Yangyi Chen, Karan Sikka, Michael Cogswell, Heng Ji, and Ajay Divakaran.
\newblock Dress: Instructing large vision-language models to align and interact with humans via natural language feedback.
\newblock {\em arXiv preprint arXiv:2311.10081}, 2023.

\bibitem{cheng2018query}
Minhao Cheng, Thong Le, Pin-Yu Chen, Jinfeng Yi, Huan Zhang, and Cho-Jui Hsieh.
\newblock Query-efficient hard-label black-box attack: An optimization-based approach.
\newblock {\em arXiv preprint arXiv:1807.04457}, 2018.

\bibitem{cheng2019sign}
Minhao Cheng, Simranjit Singh, Patrick Chen, Pin-Yu Chen, Sijia Liu, and Cho-Jui Hsieh.
\newblock Sign-opt: A query-efficient hard-label adversarial attack.
\newblock {\em arXiv preprint arXiv:1909.10773}, 2019.

\bibitem{chiang2023vicuna}
Wei-Lin Chiang, Zhuohan Li, Zi Lin, Ying Sheng, Zhanghao Wu, Hao Zhang, Lianmin Zheng, Siyuan Zhuang, Yonghao Zhuang, Joseph~E Gonzalez, et~al.
\newblock Vicuna: An open-source chatbot impressing gpt-4 with 90\%* chatgpt quality.
\newblock {\em See https://vicuna. lmsys. org (accessed 14 April 2023)}, 2(3):6, 2023.

\bibitem{chowdhury2024breaking}
Arijit~Ghosh Chowdhury, Md~Mofijul Islam, Vaibhav Kumar, Faysal~Hossain Shezan, Vinija Jain, and Aman Chadha.
\newblock Breaking down the defenses: A comparative survey of attacks on large language models.
\newblock {\em arXiv preprint arXiv:2403.04786}, 2024.

\bibitem{chung2024scaling}
Hyung~Won Chung, Le Hou, Shayne Longpre, Barret Zoph, Yi Tay, William Fedus, Yunxuan Li, Xuezhi Wang, Mostafa Dehghani, Siddhartha Brahma, et~al.
\newblock Scaling instruction-finetuned language models.
\newblock {\em Journal of Machine Learning Research}, 25(70):1--53, 2024.

\bibitem{croce2020reliable}
Francesco Croce and Matthias Hein.
\newblock Reliable evaluation of adversarial robustness with an ensemble of diverse parameter-free attacks.
\newblock In {\em International conference on machine learning}, pages 2206--2216. PMLR, 2020.

\bibitem{cui2023robustness}
Xuanimng Cui, Alejandro Aparcedo, Young~Kyun Jang, and Ser-Nam Lim.
\newblock On the robustness of large multimodal models against image adversarial attacks.
\newblock {\em arXiv preprint arXiv:2312.03777}, 2023.

\bibitem{dai2024instructblip}
Wenliang Dai, Junnan Li, Dongxu Li, Anthony Meng~Huat Tiong, Junqi Zhao, Weisheng Wang, Boyang Li, Pascale~N Fung, and Steven Hoi.
\newblock Instructblip: Towards general-purpose vision-language models with instruction tuning.
\newblock {\em Advances in Neural Information Processing Systems}, 36, 2024.

\bibitem{deng2009imagenet}
Jia Deng, Wei Dong, Richard Socher, Li-Jia Li, Kai Li, and Li Fei-Fei.
\newblock Imagenet: A large-scale hierarchical image database.
\newblock In {\em 2009 IEEE conference on computer vision and pattern recognition}, pages 248--255. Ieee, 2009.

\bibitem{dong2020survey}
Xibin Dong, Zhiwen Yu, Wenming Cao, Yifan Shi, and Qianli Ma.
\newblock A survey on ensemble learning.
\newblock {\em Frontiers of Computer Science}, 14:241--258, 2020.

\bibitem{dong2023robust}
Yinpeng Dong, Huanran Chen, Jiawei Chen, Zhengwei Fang, Xiao Yang, Yichi Zhang, Yu Tian, Hang Su, and Jun Zhu.
\newblock How robust is google's bard to adversarial image attacks?
\newblock {\em arXiv preprint arXiv:2309.11751}, 2023.

\bibitem{doshi2017towards}
Finale Doshi-Velez and Been Kim.
\newblock Towards a rigorous science of interpretable machine learning.
\newblock {\em arXiv preprint arXiv:1702.08608}, 2017.

\bibitem{fan2024unbridled}
Yihe Fan, Yuxin Cao, Ziyu Zhao, Ziyao Liu, and Shaofeng Li.
\newblock Unbridled icarus: A survey of the potential perils of image inputs in multimodal large language model security.
\newblock {\em arXiv preprint arXiv:2404.05264}, 2024.

\bibitem{fang2023eva}
Yuxin Fang, Wen Wang, Binhui Xie, Quan Sun, Ledell Wu, Xinggang Wang, Tiejun Huang, Xinlong Wang, and Yue Cao.
\newblock Eva: Exploring the limits of masked visual representation learning at scale.
\newblock In {\em Proceedings of the IEEE/CVF Conference on Computer Vision and Pattern Recognition}, pages 19358--19369, 2023.

\bibitem{fu2023misusing}
Xiaohan Fu, Zihan Wang, Shuheng Li, Rajesh~K Gupta, Niloofar Mireshghallah, Taylor Berg-Kirkpatrick, and Earlence Fernandes.
\newblock Misusing tools in large language models with visual adversarial examples.
\newblock {\em arXiv preprint arXiv:2310.03185}, 2023.

\bibitem{gao2024adversarial}
Kuofeng Gao, Yang Bai, Jiawang Bai, Yong Yang, and Shu-Tao Xia.
\newblock Adversarial robustness for visual grounding of multimodal large language models.
\newblock {\em arXiv preprint arXiv:2405.09981}, 2024.

\bibitem{gao2024inducing}
Kuofeng Gao, Yang Bai, Jindong Gu, Shu-Tao Xia, Philip Torr, Zhifeng Li, and Wei Liu.
\newblock Inducing high energy-latency of large vision-language models with verbose images.
\newblock {\em arXiv preprint arXiv:2401.11170}, 2024.

\bibitem{gao2023llama}
Peng Gao, Jiaming Han, Renrui Zhang, Ziyi Lin, Shijie Geng, Aojun Zhou, Wei Zhang, Pan Lu, Conghui He, Xiangyu Yue, et~al.
\newblock Llama-adapter v2: Parameter-efficient visual instruction model.
\newblock {\em arXiv preprint arXiv:2304.15010}, 2023.

\bibitem{gao2024sphinx}
Peng Gao, Renrui Zhang, Chris Liu, Longtian Qiu, Siyuan Huang, Weifeng Lin, Shitian Zhao, Shijie Geng, Ziyi Lin, Peng Jin, et~al.
\newblock Sphinx-x: Scaling data and parameters for a family of multi-modal large language models.
\newblock {\em arXiv preprint arXiv:2402.05935}, 2024.

\bibitem{gehman2020realtoxicityprompts}
Samuel Gehman, Suchin Gururangan, Maarten Sap, Yejin Choi, and Noah~A Smith.
\newblock Realtoxicityprompts: Evaluating neural toxic degeneration in language models.
\newblock {\em arXiv preprint arXiv:2009.11462}, 2020.

\bibitem{girdhar2023imagebind}
Rohit Girdhar, Alaaeldin El-Nouby, Zhuang Liu, Mannat Singh, Kalyan~Vasudev Alwala, Armand Joulin, and Ishan Misra.
\newblock Imagebind: One embedding space to bind them all.
\newblock In {\em Proceedings of the IEEE/CVF Conference on Computer Vision and Pattern Recognition}, pages 15180--15190, 2023.

\bibitem{gong2023figstep}
Yichen Gong, Delong Ran, Jinyuan Liu, Conglei Wang, Tianshuo Cong, Anyu Wang, Sisi Duan, and Xiaoyun Wang.
\newblock Figstep: Jailbreaking large vision-language models via typographic visual prompts.
\newblock {\em arXiv preprint arXiv:2311.05608}, 2023.

\bibitem{goodfellow2014explaining}
Ian~J Goodfellow, Jonathon Shlens, and Christian Szegedy.
\newblock Explaining and harnessing adversarial examples.
\newblock {\em arXiv preprint arXiv:1412.6572}, 2014.

\bibitem{gou2024eyes}
Yunhao Gou, Kai Chen, Zhili Liu, Lanqing Hong, Hang Xu, Zhenguo Li, Dit-Yan Yeung, James~T Kwok, and Yu Zhang.
\newblock Eyes closed, safety on: Protecting multimodal llms via image-to-text transformation.
\newblock {\em arXiv preprint arXiv:2403.09572}, 2024.

\bibitem{goyal2017making}
Yash Goyal, Tejas Khot, Douglas Summers-Stay, Dhruv Batra, and Devi Parikh.
\newblock Making the v in vqa matter: Elevating the role of image understanding in visual question answering.
\newblock In {\em Proceedings of the IEEE conference on computer vision and pattern recognition}, pages 6904--6913, 2017.

\bibitem{gu2024agent}
Xiangming Gu, Xiaosen Zheng, Tianyu Pang, Chao Du, Qian Liu, Ye Wang, Jing Jiang, and Min Lin.
\newblock Agent smith: A single image can jailbreak one million multimodal llm agents exponentially fast.
\newblock {\em arXiv preprint arXiv:2402.08567}, 2024.

\bibitem{guo2019simple}
Chuan Guo, Jacob Gardner, Yurong You, Andrew~Gordon Wilson, and Kilian Weinberger.
\newblock Simple black-box adversarial attacks.
\newblock In {\em International conference on machine learning}, pages 2484--2493. PMLR, 2019.

\bibitem{guo2024efficiently}
Qi Guo, Shanmin Pang, Xiaojun Jia, and Qing Guo.
\newblock Efficiently adversarial examples generation for visual-language models under targeted transfer scenarios using diffusion models.
\newblock {\em arXiv preprint arXiv:2404.10335}, 2024.

\bibitem{hendrycks2019benchmarking}
Dan Hendrycks and Thomas Dietterich.
\newblock Benchmarking neural network robustness to common corruptions and perturbations.
\newblock {\em arXiv preprint arXiv:1903.12261}, 2019.

\bibitem{hoffmann2022training}
Jordan Hoffmann, Sebastian Borgeaud, Arthur Mensch, Elena Buchatskaya, Trevor Cai, Eliza Rutherford, Diego de~Las Casas, Lisa~Anne Hendricks, Johannes Welbl, Aidan Clark, et~al.
\newblock Training compute-optimal large language models.
\newblock {\em arXiv preprint arXiv:2203.15556}, 2022.

\bibitem{hu2023survey}
Linmei Hu, Zeyi Liu, Ziwang Zhao, Lei Hou, Liqiang Nie, and Juanzi Li.
\newblock A survey of knowledge enhanced pre-trained language models.
\newblock {\em IEEE Transactions on Knowledge and Data Engineering}, 2023.

\bibitem{hu2024bliva}
Wenbo Hu, Yifan Xu, Yi Li, Weiyue Li, Zeyuan Chen, and Zhuowen Tu.
\newblock Bliva: A simple multimodal llm for better handling of text-rich visual questions.
\newblock In {\em Proceedings of the AAAI Conference on Artificial Intelligence}, volume~38, pages 2256--2264, 2024.

\bibitem{huang2011adversarial}
Ling Huang, Anthony~D Joseph, Blaine Nelson, Benjamin~IP Rubinstein, and J~Doug Tygar.
\newblock Adversarial machine learning.
\newblock In {\em Proceedings of the 4th ACM workshop on Security and artificial intelligence}, pages 43--58, 2011.

\bibitem{jiang2024mixtral}
Albert~Q Jiang, Alexandre Sablayrolles, Antoine Roux, Arthur Mensch, Blanche Savary, Chris Bamford, Devendra~Singh Chaplot, Diego de~las Casas, Emma~Bou Hanna, Florian Bressand, et~al.
\newblock Mixtral of experts.
\newblock {\em arXiv preprint arXiv:2401.04088}, 2024.

\bibitem{kurakin2016adversarial}
Alexey Kurakin, Ian Goodfellow, and Samy Bengio.
\newblock Adversarial machine learning at scale.
\newblock {\em arXiv preprint arXiv:1611.01236}, 2016.

\bibitem{li2023blip}
Junnan Li, Dongxu Li, Silvio Savarese, and Steven Hoi.
\newblock Blip-2: Bootstrapping language-image pre-training with frozen image encoders and large language models.
\newblock In {\em International conference on machine learning}, pages 19730--19742. PMLR, 2023.

\bibitem{li2022blip}
Junnan Li, Dongxu Li, Caiming Xiong, and Steven Hoi.
\newblock Blip: Bootstrapping language-image pre-training for unified vision-language understanding and generation.
\newblock In {\em International conference on machine learning}, pages 12888--12900. PMLR, 2022.

\bibitem{li2024empowering}
Jiatong Li, Yunqing Liu, Wenqi Fan, Xiao-Yong Wei, Hui Liu, Jiliang Tang, and Qing Li.
\newblock Empowering molecule discovery for molecule-caption translation with large language models: A chatgpt perspective.
\newblock {\em IEEE Transactions on Knowledge and Data Engineering}, 2024.

\bibitem{li2024one}
Lin Li, Haoyan Guan, Jianing Qiu, and Michael Spratling.
\newblock One prompt word is enough to boost adversarial robustness for pre-trained vision-language models.
\newblock {\em arXiv preprint arXiv:2403.01849}, 2024.

\bibitem{li2024red}
Mukai Li, Lei Li, Yuwei Yin, Masood Ahmed, Zhenguang Liu, and Qi Liu.
\newblock Red teaming visual language models.
\newblock {\em arXiv preprint arXiv:2401.12915}, 2024.

\bibitem{li2024images}
Yifan Li, Hangyu Guo, Kun Zhou, Wayne~Xin Zhao, and Ji-Rong Wen.
\newblock Images are achilles' heel of alignment: Exploiting visual vulnerabilities for jailbreaking multimodal large language models.
\newblock {\em arXiv preprint arXiv:2403.09792}, 2024.

\bibitem{liang2024vl}
Jiawei Liang, Siyuan Liang, Man Luo, Aishan Liu, Dongchen Han, Ee-Chien Chang, and Xiaochun Cao.
\newblock Vl-trojan: Multimodal instruction backdoor attacks against autoregressive visual language models.
\newblock {\em arXiv preprint arXiv:2402.13851}, 2024.

\bibitem{liang2019efficient}
Jinwen Liang, Zheng Qin, Sheng Xiao, Lu Ou, and Xiaodong Lin.
\newblock Efficient and secure decision tree classification for cloud-assisted online diagnosis services.
\newblock {\em IEEE Transactions on Dependable and Secure Computing}, 18(4):1632--1644, 2019.

\bibitem{liang2024revisiting}
Siyuan Liang, Jiawei Liang, Tianyu Pang, Chao Du, Aishan Liu, Ee-Chien Chang, and Xiaochun Cao.
\newblock Revisiting backdoor attacks against large vision-language models.
\newblock {\em arXiv preprint arXiv:2406.18844}, 2024.

\bibitem{lin2014microsoft}
Tsung-Yi Lin, Michael Maire, Serge Belongie, James Hays, Pietro Perona, Deva Ramanan, Piotr Doll{\'a}r, and C~Lawrence Zitnick.
\newblock Microsoft coco: Common objects in context.
\newblock In {\em Computer Vision--ECCV 2014: 13th European Conference, Zurich, Switzerland, September 6-12, 2014, Proceedings, Part V 13}, pages 740--755. Springer, 2014.

\bibitem{lin2023pre}
Yan Lin, Huaiyu Wan, Shengnan Guo, Jilin Hu, Christian~S Jensen, and Youfang Lin.
\newblock Pre-training general trajectory embeddings with maximum multi-view entropy coding.
\newblock {\em IEEE Transactions on Knowledge and Data Engineering}, 2023.

\bibitem{lin2023toxicchat}
Zi Lin, Zihan Wang, Yongqi Tong, Yangkun Wang, Yuxin Guo, Yujia Wang, and Jingbo Shang.
\newblock Toxicchat: Unveiling hidden challenges of toxicity detection in real-world user-ai conversation.
\newblock {\em arXiv preprint arXiv:2310.17389}, 2023.

\bibitem{liu2022imperceptible}
Daizong Liu and Wei Hu.
\newblock Imperceptible transfer attack and defense on 3d point cloud classification.
\newblock {\em IEEE transactions on pattern analysis and machine intelligence}, 45(4):4727--4746, 2022.

\bibitem{liu2024improved}
Haotian Liu, Chunyuan Li, Yuheng Li, and Yong~Jae Lee.
\newblock Improved baselines with visual instruction tuning.
\newblock In {\em Proceedings of the IEEE/CVF Conference on Computer Vision and Pattern Recognition}, pages 26296--26306, 2024.

\bibitem{liu2024visual}
Haotian Liu, Chunyuan Li, Qingyang Wu, and Yong~Jae Lee.
\newblock Visual instruction tuning.
\newblock {\em Advances in neural information processing systems}, 36, 2024.

\bibitem{liu2024mmsafetybench}
Xin Liu, Yichen Zhu, Jindong Gu, Yunshi Lan, Chao Yang, and Yu Qiao.
\newblock Mm-safetybench: A benchmark for safety evaluation of multimodal large language models, 2024.

\bibitem{liu2024safety}
Xin Liu, Yichen Zhu, Yunshi Lan, Chao Yang, and Yu Qiao.
\newblock Safety of multimodal large language models on images and text.
\newblock {\em arXiv preprint arXiv:2402.00357}, 2024.

\bibitem{liu2016delving}
Yanpei Liu, Xinyun Chen, Chang Liu, and Dawn Song.
\newblock Delving into transferable adversarial examples and black-box attacks.
\newblock {\em arXiv preprint arXiv:1611.02770}, 2016.

\bibitem{lu2024test}
Dong Lu, Tianyu Pang, Chao Du, Qian Liu, Xianjun Yang, and Min Lin.
\newblock Test-time backdoor attacks on multimodal large language models.
\newblock {\em arXiv preprint arXiv:2402.08577}, 2024.

\bibitem{luo2024image}
Haochen Luo, Jindong Gu, Fengyuan Liu, and Philip Torr.
\newblock An image is worth 1000 lies: Adversarial transferability across prompts on vision-language models.
\newblock {\em arXiv preprint arXiv:2403.09766}, 2024.

\bibitem{luo2024jailbreakv}
Weidi Luo, Siyuan Ma, Xiaogeng Liu, Xiaoyu Guo, and Chaowei Xiao.
\newblock Jailbreakv-28k: A benchmark for assessing the robustness of multimodal large language models against jailbreak attacks.
\newblock {\em arXiv preprint arXiv:2404.03027}, 2024.

\bibitem{ma2024visual}
Siyuan Ma, Weidi Luo, Yu Wang, Xiaogeng Liu, Muhao Chen, Bo Li, and Chaowei Xiao.
\newblock Visual-roleplay: Universal jailbreak attack on multimodal large language models via role-playing image characte.
\newblock {\em arXiv preprint arXiv:2405.20773}, 2024.

\bibitem{madry2017towards}
Aleksander Madry, Aleksandar Makelov, Ludwig Schmidt, Dimitris Tsipras, and Adrian Vladu.
\newblock Towards deep learning models resistant to adversarial attacks.
\newblock {\em arXiv preprint arXiv:1706.06083}, 2017.

\bibitem{marino2019ok}
Kenneth Marino, Mohammad Rastegari, Ali Farhadi, and Roozbeh Mottaghi.
\newblock Ok-vqa: A visual question answering benchmark requiring external knowledge.
\newblock In {\em Proceedings of the IEEE/cvf conference on computer vision and pattern recognition}, pages 3195--3204, 2019.

\bibitem{mehrabi2021survey}
Ninareh Mehrabi, Fred Morstatter, Nripsuta Saxena, Kristina Lerman, and Aram Galstyan.
\newblock A survey on bias and fairness in machine learning.
\newblock {\em ACM computing surveys (CSUR)}, 54(6):1--35, 2021.

\bibitem{mitnick2003art}
Kevin~D Mitnick and William~L Simon.
\newblock {\em The art of deception: Controlling the human element of security}.
\newblock John Wiley \& Sons, 2003.

\bibitem{moosavi2017universal}
Seyed-Mohsen Moosavi-Dezfooli, Alhussein Fawzi, Omar Fawzi, and Pascal Frossard.
\newblock Universal adversarial perturbations.
\newblock In {\em Proceedings of the IEEE conference on computer vision and pattern recognition}, pages 1765--1773, 2017.

\bibitem{muhammad2020deep}
Khan Muhammad, Amin Ullah, Jaime Lloret, Javier Del~Ser, and Victor Hugo~C de Albuquerque.
\newblock Deep learning for safe autonomous driving: Current challenges and future directions.
\newblock {\em IEEE Transactions on Intelligent Transportation Systems}, 22(7):4316--4336, 2020.

\bibitem{nesterov2017random}
Yurii Nesterov and Vladimir Spokoiny.
\newblock Random gradient-free minimization of convex functions.
\newblock {\em Foundations of Computational Mathematics}, 17(2):527--566, 2017.

\bibitem{ni2024physical}
Zhenyang Ni, Rui Ye, Yuxi Wei, Zhen Xiang, Yanfeng Wang, and Siheng Chen.
\newblock Physical backdoor attack can jeopardize driving with vision-large-language models.
\newblock {\em arXiv preprint arXiv:2404.12916}, 2024.

\bibitem{nichol2021glide}
Alex Nichol, Prafulla Dhariwal, Aditya Ramesh, Pranav Shyam, Pamela Mishkin, Bob McGrew, Ilya Sutskever, and Mark Chen.
\newblock Glide: Towards photorealistic image generation and editing with text-guided diffusion models.
\newblock {\em arXiv preprint arXiv:2112.10741}, 2021.

\bibitem{niu2024jailbreaking}
Zhenxing Niu, Haodong Ren, Xinbo Gao, Gang Hua, and Rong Jin.
\newblock Jailbreaking attack against multimodal large language model.
\newblock {\em arXiv preprint arXiv:2402.02309}, 2024.

\bibitem{OpenAI2023card}
OpenAI.
\newblock Gpt-4v(ision) system card.
\newblock 2023.

\bibitem{pan2024unifying}
Shirui Pan, Linhao Luo, Yufei Wang, Chen Chen, Jiapu Wang, and Xindong Wu.
\newblock Unifying large language models and knowledge graphs: A roadmap.
\newblock {\em IEEE Transactions on Knowledge and Data Engineering}, 2024.

\bibitem{pi2024mllm}
Renjie Pi, Tianyang Han, Yueqi Xie, Rui Pan, Qing Lian, Hanze Dong, Jipeng Zhang, and Tong Zhang.
\newblock Mllm-protector: Ensuring mllm's safety without hurting performance.
\newblock {\em arXiv preprint arXiv:2401.02906}, 2024.

\bibitem{plummer2015flickr30k}
Bryan~A Plummer, Liwei Wang, Chris~M Cervantes, Juan~C Caicedo, Julia Hockenmaier, and Svetlana Lazebnik.
\newblock Flickr30k entities: Collecting region-to-phrase correspondences for richer image-to-sentence models.
\newblock In {\em Proceedings of the IEEE international conference on computer vision}, pages 2641--2649, 2015.

\bibitem{polikar2012ensemble}
Robi Polikar.
\newblock Ensemble learning.
\newblock {\em Ensemble machine learning: Methods and applications}, pages 1--34, 2012.

\bibitem{qi2024visual}
Xiangyu Qi, Kaixuan Huang, Ashwinee Panda, Peter Henderson, Mengdi Wang, and Prateek Mittal.
\newblock Visual adversarial examples jailbreak aligned large language models.
\newblock In {\em Proceedings of the AAAI Conference on Artificial Intelligence}, volume~38, pages 21527--21536, 2024.

\bibitem{qraitem2024vision}
Maan Qraitem, Nazia Tasnim, Kate Saenko, and Bryan~A Plummer.
\newblock Vision-llms can fool themselves with self-generated typographic attacks.
\newblock {\em arXiv preprint arXiv:2402.00626}, 2024.

\bibitem{radford2021learning}
Alec Radford, Jong~Wook Kim, Chris Hallacy, Aditya Ramesh, Gabriel Goh, Sandhini Agarwal, Girish Sastry, Amanda Askell, Pamela Mishkin, Jack Clark, et~al.
\newblock Learning transferable visual models from natural language supervision.
\newblock In {\em International conference on machine learning}, pages 8748--8763. PMLR, 2021.

\bibitem{radford2019language}
Alec Radford, Jeffrey Wu, Rewon Child, David Luan, Dario Amodei, Ilya Sutskever, et~al.
\newblock Language models are unsupervised multitask learners.
\newblock {\em OpenAI blog}, 1(8):9, 2019.

\bibitem{ramesh2022hierarchical}
Aditya Ramesh, Prafulla Dhariwal, Alex Nichol, Casey Chu, and Mark Chen.
\newblock Hierarchical text-conditional image generation with clip latents.
\newblock {\em arXiv preprint arXiv:2204.06125}, 1(2):3, 2022.

\bibitem{ramesh2021zero}
Aditya Ramesh, Mikhail Pavlov, Gabriel Goh, Scott Gray, Chelsea Voss, Alec Radford, Mark Chen, and Ilya Sutskever.
\newblock Zero-shot text-to-image generation.
\newblock In {\em International conference on machine learning}, pages 8821--8831. Pmlr, 2021.

\bibitem{rombach2022high}
Robin Rombach, Andreas Blattmann, Dominik Lorenz, Patrick Esser, and Bj{\"o}rn Ommer.
\newblock High-resolution image synthesis with latent diffusion models.
\newblock In {\em Proceedings of the IEEE/CVF conference on computer vision and pattern recognition}, pages 10684--10695, 2022.

\bibitem{schlarmann2023adversarial}
Christian Schlarmann and Matthias Hein.
\newblock On the adversarial robustness of multi-modal foundation models.
\newblock In {\em Proceedings of the IEEE/CVF International Conference on Computer Vision}, pages 3677--3685, 2023.

\bibitem{shalev2016safe}
Shai Shalev-Shwartz, Shaked Shammah, and Amnon Shashua.
\newblock Safe, multi-agent, reinforcement learning for autonomous driving.
\newblock {\em arXiv preprint arXiv:1610.03295}, 2016.

\bibitem{shayegani2023jailbreak}
Erfan Shayegani, Yue Dong, and Nael Abu-Ghazaleh.
\newblock Jailbreak in pieces: Compositional adversarial attacks on multi-modal language models.
\newblock In {\em The Twelfth International Conference on Learning Representations}, 2023.

\bibitem{su2023pandagpt}
Yixuan Su, Tian Lan, Huayang Li, Jialu Xu, Yan Wang, and Deng Cai.
\newblock Pandagpt: One model to instruction-follow them all.
\newblock {\em arXiv preprint arXiv:2305.16355}, 2023.

\bibitem{tan2024wolf}
Zhen Tan, Chengshuai Zhao, Raha Moraffah, Yifan Li, Yu Kong, Tianlong Chen, and Huan Liu.
\newblock The wolf within: Covert injection of malice into mllm societies via an mllm operative.
\newblock {\em arXiv preprint arXiv:2402.14859}, 2024.

\bibitem{tao2024imgtrojan}
Xijia Tao, Shuai Zhong, Lei Li, Qi Liu, and Lingpeng Kong.
\newblock Imgtrojan: Jailbreaking vision-language models with one image.
\newblock {\em arXiv preprint arXiv:2403.02910}, 2024.

\bibitem{mosaicml2023introducing}
MosaicML~NLP Team et~al.
\newblock Introducing mpt-7b: A new standard for open-source, commercially usable llms, 2023.
\newblock {\em URL www. mosaicml. com/blog/mpt-7b. Accessed}, pages 05--05, 2023.

\bibitem{touvron2023llama}
Hugo Touvron, Thibaut Lavril, Gautier Izacard, Xavier Martinet, Marie-Anne Lachaux, Timoth{\'e}e Lacroix, Baptiste Rozi{\`e}re, Naman Goyal, Eric Hambro, Faisal Azhar, et~al.
\newblock Llama: Open and efficient foundation language models.
\newblock {\em arXiv preprint arXiv:2302.13971}, 2023.

\bibitem{touvron2023llama2}
Hugo Touvron, Louis Martin, Kevin Stone, Peter Albert, Amjad Almahairi, Yasmine Babaei, Nikolay Bashlykov, Soumya Batra, Prajjwal Bhargava, Shruti Bhosale, et~al.
\newblock Llama 2: Open foundation and fine-tuned chat models.
\newblock {\em arXiv preprint arXiv:2307.09288}, 2023.

\bibitem{tramer2017ensemble}
Florian Tram{\`e}r, Alexey Kurakin, Nicolas Papernot, Ian Goodfellow, Dan Boneh, and Patrick McDaniel.
\newblock Ensemble adversarial training: Attacks and defenses.
\newblock {\em arXiv preprint arXiv:1705.07204}, 2017.

\bibitem{tsimpoukelli2021multimodal}
Maria Tsimpoukelli, Jacob~L Menick, Serkan Cabi, SM Eslami, Oriol Vinyals, and Felix Hill.
\newblock Multimodal few-shot learning with frozen language models.
\newblock {\em Advances in Neural Information Processing Systems}, 34:200--212, 2021.

\bibitem{tu2023many}
Haoqin Tu, Chenhang Cui, Zijun Wang, Yiyang Zhou, Bingchen Zhao, Junlin Han, Wangchunshu Zhou, Huaxiu Yao, and Cihang Xie.
\newblock How many unicorns are in this image? a safety evaluation benchmark for vision llms.
\newblock {\em arXiv preprint arXiv:2311.16101}, 2023.

\bibitem{vaswani2017attention}
Ashish Vaswani, Noam Shazeer, Niki Parmar, Jakob Uszkoreit, Llion Jones, Aidan~N Gomez, {\L}ukasz Kaiser, and Illia Polosukhin.
\newblock Attention is all you need.
\newblock {\em Advances in neural information processing systems}, 30, 2017.

\bibitem{walmer2022dual}
Matthew Walmer, Karan Sikka, Indranil Sur, Abhinav Shrivastava, and Susmit Jha.
\newblock Dual-key multimodal backdoors for visual question answering.
\newblock In {\em Proceedings of the IEEE/CVF Conference on computer vision and pattern recognition}, pages 15375--15385, 2022.

\bibitem{wang2024transferable}
Haodi Wang, Kai Dong, Zhilei Zhu, Haotong Qin, Aishan Liu, Xiaolin Fang, Jiakai Wang, and Xianglong Liu.
\newblock Transferable multimodal attack on vision-language pre-training models.
\newblock In {\em 2024 IEEE Symposium on Security and Privacy (SP)}, pages 102--102. IEEE Computer Society, 2024.

\bibitem{wang2024inferaligner}
Pengyu Wang, Dong Zhang, Linyang Li, Chenkun Tan, Xinghao Wang, Ke Ren, Botian Jiang, and Xipeng Qiu.
\newblock Inferaligner: Inference-time alignment for harmlessness through cross-model guidance.
\newblock {\em arXiv preprint arXiv:2401.11206}, 2024.

\bibitem{wang2024white}
Ruofan Wang, Xingjun Ma, Hanxu Zhou, Chuanjun Ji, Guangnan Ye, and Yu-Gang Jiang.
\newblock White-box multimodal jailbreaks against large vision-language models.
\newblock {\em arXiv preprint arXiv:2405.17894}, 2024.

\bibitem{wang2024llms}
Siyuan Wang, Zhuohan Long, Zhihao Fan, and Zhongyu Wei.
\newblock From llms to mllms: Exploring the landscape of multimodal jailbreaking.
\newblock {\em arXiv preprint arXiv:2406.14859}, 2024.

\bibitem{wang2021enhancing}
Xiaosen Wang and Kun He.
\newblock Enhancing the transferability of adversarial attacks through variance tuning.
\newblock In {\em Proceedings of the IEEE/CVF conference on computer vision and pattern recognition}, pages 1924--1933, 2021.

\bibitem{wang2023instructta}
Xunguang Wang, Zhenlan Ji, Pingchuan Ma, Zongjie Li, and Shuai Wang.
\newblock Instructta: Instruction-tuned targeted attack for large vision-language models.
\newblock {\em arXiv preprint arXiv:2312.01886}, 2023.

\bibitem{wang2019transferable}
Ximei Wang, Ying Jin, Mingsheng Long, Jianmin Wang, and Michael~I Jordan.
\newblock Transferable normalization: Towards improving transferability of deep neural networks.
\newblock {\em Advances in neural information processing systems}, 32, 2019.

\bibitem{wang2024adashield}
Yu Wang, Xiaogeng Liu, Yu Li, Muhao Chen, and Chaowei Xiao.
\newblock Adashield: Safeguarding multimodal large language models from structure-based attack via adaptive shield prompting.
\newblock {\em arXiv preprint arXiv:2403.09513}, 2024.

\bibitem{wang2024stop}
Zefeng Wang, Zhen Han, Shuo Chen, Fan Xue, Zifeng Ding, Xun Xiao, Volker Tresp, Philip Torr, and Jindong Gu.
\newblock Stop reasoning! when multimodal llms with chain-of-thought reasoning meets adversarial images.
\newblock {\em arXiv preprint arXiv:2402.14899}, 2024.

\bibitem{midjourney}
Midjourney website.
\newblock Midjourney, 2023.

\bibitem{wu2024adversarial}
Chen~Henry Wu, Jing~Yu Koh, Ruslan Salakhutdinov, Daniel Fried, and Aditi Raghunathan.
\newblock Adversarial attacks on multimodal agents.
\newblock {\em arXiv preprint arXiv:2406.12814}, 2024.

\bibitem{wu2023jailbreaking}
Yuanwei Wu, Xiang Li, Yixin Liu, Pan Zhou, and Lichao Sun.
\newblock Jailbreaking gpt-4v via self-adversarial attacks with system prompts.
\newblock {\em arXiv preprint arXiv:2311.09127}, 2023.

\bibitem{xu2024shadowcast}
Yuancheng Xu, Jiarui Yao, Manli Shu, Yanchao Sun, Zichu Wu, Ning Yu, Tom Goldstein, and Furong Huang.
\newblock Shadowcast: Stealthy data poisoning attacks against vision-language models.
\newblock {\em arXiv preprint arXiv:2402.06659}, 2024.

\bibitem{yang2024robust}
Wenhan Yang, Jingdong Gao, and Baharan Mirzasoleiman.
\newblock Robust contrastive language-image pretraining against data poisoning and backdoor attacks.
\newblock {\em Advances in Neural Information Processing Systems}, 36, 2024.

\bibitem{yang2023dawn}
Zhengyuan Yang, Linjie Li, Kevin Lin, Jianfeng Wang, Chung-Ching Lin, Zicheng Liu, and Lijuan Wang.
\newblock The dawn of lmms: Preliminary explorations with gpt-4v (ision).
\newblock {\em arXiv preprint arXiv:2309.17421}, 9(1):1, 2023.

\bibitem{ying2024unveiling}
Zonghao Ying, Aishan Liu, Xianglong Liu, and Dacheng Tao.
\newblock Unveiling the safety of gpt-4o: An empirical study using jailbreak attacks.
\newblock {\em arXiv preprint arXiv:2406.06302}, 2024.

\bibitem{yurtsever2020survey}
Ekim Yurtsever, Jacob Lambert, Alexander Carballo, and Kazuya Takeda.
\newblock A survey of autonomous driving: Common practices and emerging technologies.
\newblock {\em IEEE access}, 8:58443--58469, 2020.

\bibitem{zhang2024avibench}
Hao Zhang, Wenqi Shao, Hong Liu, Yongqiang Ma, Ping Luo, Yu Qiao, and Kaipeng Zhang.
\newblock Avibench: Towards evaluating the robustness of large vision-language model on adversarial visual-instructions.
\newblock {\em arXiv preprint arXiv:2403.09346}, 2024.

\bibitem{zhang2023adversarial}
Jiaming Zhang, Xingjun Ma, Xin Wang, Lingyu Qiu, Jiaqi Wang, Yu-Gang Jiang, and Jitao Sang.
\newblock Adversarial prompt tuning for vision-language models.
\newblock {\em arXiv preprint arXiv:2311.11261}, 2023.

\bibitem{zhang2022improving}
Jianping Zhang, Weibin Wu, Jen-tse Huang, Yizhan Huang, Wenxuan Wang, Yuxin Su, and Michael~R Lyu.
\newblock Improving adversarial transferability via neuron attribution-based attacks.
\newblock In {\em Proceedings of the IEEE/CVF Conference on Computer Vision and Pattern Recognition}, pages 14993--15002, 2022.

\bibitem{zhang2022opt}
Susan Zhang, Stephen Roller, Naman Goyal, Mikel Artetxe, Moya Chen, Shuohui Chen, Christopher Dewan, Mona Diab, Xian Li, Xi~Victoria Lin, et~al.
\newblock Opt: Open pre-trained transformer language models.
\newblock {\em arXiv preprint arXiv:2205.01068}, 2022.

\bibitem{zhang2023mutation}
Xiaoyu Zhang, Cen Zhang, Tianlin Li, Yihao Huang, Xiaojun Jia, Xiaofei Xie, Yang Liu, and Chao Shen.
\newblock A mutation-based method for multi-modal jailbreaking attack detection.
\newblock {\em arXiv preprint arXiv:2312.10766}, 2023.

\bibitem{zhao2017men}
Jieyu Zhao, Tianlu Wang, Mark Yatskar, Vicente Ordonez, and Kai-Wei Chang.
\newblock Men also like shopping: Reducing gender bias amplification using corpus-level constraints.
\newblock {\em arXiv preprint arXiv:1707.09457}, 2017.

\bibitem{zhao2024evaluating}
Yunqing Zhao, Tianyu Pang, Chao Du, Xiao Yang, Chongxuan Li, Ngai-Man~Man Cheung, and Min Lin.
\newblock On evaluating adversarial robustness of large vision-language models.
\newblock {\em Advances in Neural Information Processing Systems}, 36, 2024.

\bibitem{zhao2024recommender}
Zihuai Zhao, Wenqi Fan, Jiatong Li, Yunqing Liu, Xiaowei Mei, Yiqi Wang, Zhen Wen, Fei Wang, Xiangyu Zhao, Jiliang Tang, et~al.
\newblock Recommender systems in the era of large language models (llms).
\newblock {\em IEEE Transactions on Knowledge and Data Engineering}, 2024.

\bibitem{zhu2023minigpt}
Deyao Zhu, Jun Chen, Xiaoqian Shen, Xiang Li, and Mohamed Elhoseiny.
\newblock Minigpt-4: Enhancing vision-language understanding with advanced large language models.
\newblock {\em arXiv preprint arXiv:2304.10592}, 2023.

\bibitem{zou2023universal}
Andy Zou, Zifan Wang, J~Zico Kolter, and Matt Fredrikson.
\newblock Universal and transferable adversarial attacks on aligned language models.
\newblock {\em arXiv preprint arXiv:2307.15043}, 2023.

\end{thebibliography}

\end{document}